\documentclass[journal]{IEEEtran}
\usepackage{cite}
\usepackage{amsmath}
\usepackage{algorithmic}
\usepackage{array}
\usepackage{etoolbox}

\usepackage{url}

\usepackage{wrapfig}

\usepackage{overpic}
\usepackage{multirow}

\usepackage[latin1]{inputenc}
\usepackage{setspace}
\usepackage{fancyhdr}
\usepackage[usenames]{color}
\RequirePackage[OT1]{fontenc}
\usepackage{stmaryrd}
\usepackage{amsmath,amsfonts,amssymb,amsthm}
\usepackage{graphicx}

\usepackage{paralist}
\usepackage{amsbsy}

\usepackage[mathscr]{eucal}

\usepackage{bm}

\usepackage{xspace}

\usepackage{algorithm,algorithmic}
\mathchardef\mhyphen="2D

\newcommand{\Real}{\mathbb{R}}

\newcommand{\Tra}{^{\sf T}} % Transpose
\newcommand{\MI}[1]{\mathbf{#1}}

\newcommand{\V}[1]{{\bm{\mathbf{\MakeLowercase{#1}}}}} % vector
\newcommand{\Vtilde}[1]{{\bm{\tilde \mathbf{\MakeLowercase{#1}}}}} % vector
\newcommand{\Vn}[2]{\V{#1}^{(#2)}} % n-th vector
\newcommand{\M}[1]{{\bm{\mathbf{\MakeUppercase{#1}}}}} % matrix
\newcommand{\Mn}[2]{\M{#1}^{(#2)}} % n-th matrix
\newcommand{\MnC}[3]{\V{#1}^{(#2)}_{#3}} % n-th matrix column
\newcommand{\T}[1]{\boldsymbol{\mathscr{\MakeUppercase{#1}}}} %tensor
\newcommand{\That}[1]{\boldsymbol{\hat \mathscr{\MakeUppercase{#1}}}} %tensor
\newcommand{\TE}[2]{\MakeLowercase{#1}_{\MI{#2}}} % tensor element with multi-index
\newcommand{\KT}[1]{\left\llbracket #1 \right\rrbracket} % kruskal operator

\newcommand{\Kron}{\otimes} %Kronecker
\newcommand{\Khat}{\odot} %Khatri-Rao
\newcommand{\Mz}[2]{\M{#1}_{(#2)}} % n-th mode matricize
\newcommand{\qtext}[1]{\quad\text{#1}\quad}
\newcommand{\sign}[1]{\operatorname{sign}({#1})}

\begin{document}

\title{Shape Constrained Tensor Decompositions using Sparse Representations in Over-Complete Libraries}

\author{Bethany Lusch, Eric C. Chi,
        and~J. Nathan Kutz,~\IEEEmembership{Member,~IEEE}
\thanks{B. Lusch and J. N. Kutz are with the Department
of Applied Mathematics, University of Washington, Seattle, WA  98195-3925  USA e-mail:  herwaldt@uw.edu, kutz@uw.edu.}% <-this % stops a space
\thanks{E. Chi is with the Department of Statistics, North Carolina State University, Raleigh, NC 27695-8203 USA e-mail:  eric\_chi@ncsu.edu.}}% <-this % stops a space
% <-this % stops a space


\maketitle

\begin{abstract}
We consider $N$-way data arrays and low-rank tensor factorizations where the time mode is coded as a sparse linear combination of temporal elements from an over-complete library.  Our method, Shape Constrained Tensor Decomposition (SCTD) is based upon the CANDECOMP/PARAFAC (CP) decomposition which produces $r$-rank approximations of data tensors via outer products of vectors in
each dimension of the data.  By constraining the vector in the temporal dimension to known analytic forms which are selected from a large set of candidate functions, more readily interpretable decompositions are achieved and analytic time dependencies discovered.  The SCTD method circumvents traditional {\em flattening} techniques where an $N$-way array is reshaped into a matrix in order to perform a singular value decomposition.  A clear advantage of the SCTD algorithm is its ability to extract transient and intermittent phenomena which is often difficult for SVD-based methods.   We motivate the SCTD method using several intuitively appealing results before applying it on a number of high-dimensional, real-world data sets in order to illustrate the efficiency of the algorithm in extracting interpretable spatio-temporal modes.  With the rise of data-driven discovery methods, the decomposition proposed provides a viable technique for analyzing multitudes of data in a more comprehensible fashion.
\end{abstract}


\IEEEpeerreviewmaketitle

\section{Introduction}

\IEEEPARstart{M}{atrix} decompositions are critically enabling algorithms for scientific computing and data analysis applications across every field of the engineering, social, biological, and physical sciences.  Of particular importance is the singular value decomposition (SVD), which provides a principled method for dimensionality reduction and computation of interpretable subspaces within which the data reside.  So widespread is the usage of the algorithm, and minor modifications thereof, that it has generated a myriad of names across various communities, including Principal Component Analysis (PCA)~\cite{Pearson:1901}, the Karhunen-Lo\`eve (KL) decomposition, Hotelling transform~\cite{hotellingJEdPsy33_1,hotellingJEdPsy33_2}, Empirical Orthogonal Functions (EOFs)~\cite{lorenzMITTR56} and Proper Orthogonal Decomposition (POD)~\cite{Lumley:1970,HLBR_turb}.  However, in order to use the SVD, data, which generally may be of $N$ distinct dimensions, must be {\em flattened} into a matrix
form, potentially compromising the statistical accuracy of the subspaces computed.  Tensor decompositions are a generalization of the SVD concept to higher dimensions, allowing for $N$-way arrays ($N\geq 3$) of data to be decomposed into their constitutive, low-rank subspaces without flattening, which is especially advantageous for categorical data types.  It is often the case that one of the dimensions considered in the tensor is the time variable.   In this paper, we develop a version of a tensor decomposition algorithm that restricts the time dynamics to analytically tractable solutions sparsely selected from a large, over-complete library of candidate functions.  In so doing, we provide a more interpretable framework for the tensor modes in the decomposition process and analytic expressions for their associated time dynamics.  

With the rise of data science and data-driven discovery, tensor decompositions are of increasing value and importance for characterizing underlying structure and dimensionality of data~\cite{KolBad2009}.  Indeed, finding low-rank structure in high-dimensional data is at the core of many machine learning architectures~\cite{Bishop,Murphy}.   In applications, one of the important dimensions of the data set is a time variable which measures how the other quantities of interest evolve over a prescribed time course. A tensor decomposition produces the low-rank time variable evolution.   However, the low-rank time modes often are complicated and noisy due to the form of the data itself.  In contrast, we often expect simple and highly structured temporal signatures, whether it be oscillations of a prescribed frequency or exponential growth/decay of a signal, for instance.  The natural remedy is to constrain the form of the temporal modes extracted from the tensor decomposition.  By specifying an over-complete library of temporal functions, we are able to extract analytic forms for the best fit temporal evolution of the data.  The appropriate time behavior in our over-complete library is selected through sparse $\ell_1$ regression techniques so as to select a minimal, but most informative, set of time dynamics.  This is highly advantageous for characterizing the structure of the data and for data-driven discovery of underlying processes responsible for producing the dynamics observed.  

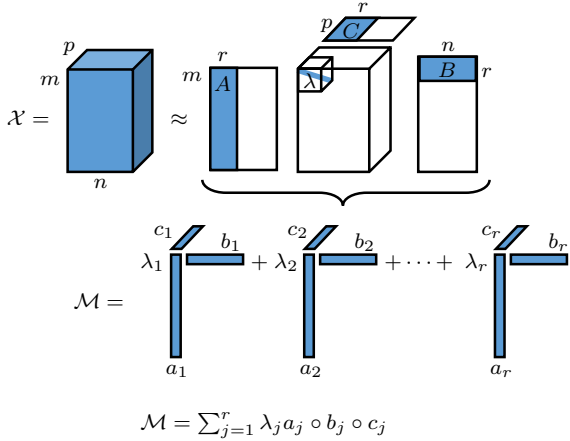
\begin{figure}[t]
\centering
\begin{overpic}[width=3.5in]{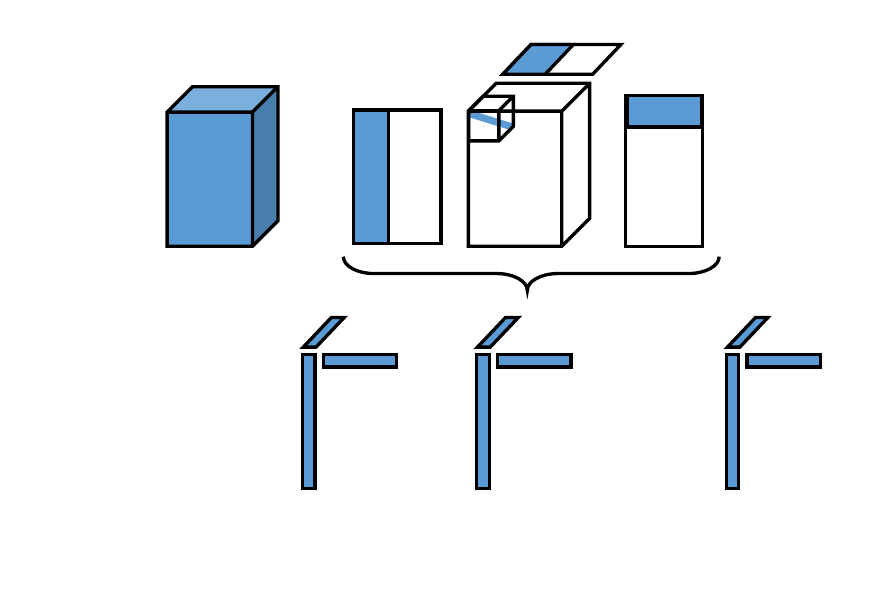}
%\begin{overpic}[width=3.5in,grid,tics=5]{./Figures/Figure3TensorDecomp2.pdf}
\put(10,47.5){{\footnotesize $\mathcal{X} = $}}
\put(15,53.5){{\footnotesize $m$}}
\put(18.5,58.5){{\footnotesize $p$}}
\put(23,38){{\footnotesize $n$}}
\put(34.5,47.5){{\footnotesize $\approx$}}
\put(36.2,54){{\footnotesize $m$}}
\put(41.5,57){{\footnotesize $r$}}
\put(41,52.5){{\footnotesize $A$}}
\put(54.3,53){{\footnotesize $\lambda$}}
\put(57,62){{\footnotesize $p$}}
\put(60.2,60.4){{\footnotesize $C$}}
\put(62.5,64.3){{\footnotesize $r$}}
\put(81,54.7){{\footnotesize $r$}}
\put(74.5,54.7){{\footnotesize $B$}}
\put(75,58.5){{\footnotesize $n$}}
\put(20,20){{\footnotesize $\mathcal{M} = $}}
\put(33.8,10.2){{\footnotesize $a_1$}}
\put(42,29){{\footnotesize $b_1$}}
\put(32,31){{\footnotesize $c_1$}}
\put(30,26){{\footnotesize $\lambda_1$}}
\put(53.8,10.2){{\footnotesize $a_2$}}
\put(62,29){{\footnotesize $b_2$}}
\put(52,31){{\footnotesize $c_2$}}
\put(46.5,26){{\footnotesize $+$}}
\put(50,26){{\footnotesize $\lambda_2$}}
\put(82.3,10.2){{\footnotesize $a_r$}}
\put(91,29){{\footnotesize $b_r$}}
\put(81,31){{\footnotesize $c_r$}}
\put(66.5,26){{\footnotesize $+\dots +$}}
\put(78.5,26){{\footnotesize $\lambda_r$}}
\put(30,2){{\footnotesize $\mathcal{M} = \sum_{j=1}^r \lambda_j a_j \circ b_j \circ c_j$}}
\end{overpic}
\caption{CP tensor decompositions. This type of decomposition approximates a data set $\mathcal{X}$ with a tensor $\mathcal{M}$ consisting of $r$ components. Each component is an outer product of three vectors and is of the form $\lambda_j a_j \circ b_j \circ c_j$.}
\label{fig:TensorDecomp}
\end{figure}

Unlike the SVD for matrices, tensor decompositions are not unique, and there are a variety of decompositions that can be applied to $N$-way arrays ($N\geq 3$) of data.  We consider the CANDECOMP/PARAFAC (CP)  decomposition~\cite{CaCh70,Ha70} illustrated in Fig.~\ref{fig:TensorDecomp}, which arranges $r$-rank data into a series of outer products of $N$ vectors. 
There are other decompositions available, including the Tucker tensor decomposition \cite{KolBad2009} and the recently developed tensor-based method \cite{tensorDMD} for Dynamic
Mode Decomposition (DMD) \cite{TuRowLuc2014}.  The former method is widely used in the tensor community while the latter method provides a regression that enforces Fourier mode behavior in the time mode.  All three methods fall short of our primary goal, which is to provide a tensor decomposition with analytically tractable time dynamics capable of modeling transient phenomena.  The DMD algorithm solves the first part of this objective but fails in modeling transient phenomena.   Although a multi-resolution DMD
method has been proposed to handle transients~\cite{mrDMD}, it has a multiple pass architecture that sometimes struggles to extract spatio-temporal structures in a completely unsupervised manner.   In this work, we demonstrate that the CP tensor decomposition can be modified to constrain the time dynamic mode to a broad range of analytic solutions that are selected from a large and over-complete library of 
candidate functions.  By using sparse regression techniques, the best candidate functions are selected
for representing the temporal dynamics.  We call this technique Shape Constrained Tensor Decomposition (SCTD). The clear advantage of the SCTD over standard CP decompositions
is that it gives analytic results which are readily interpretable.  We demonstrate the method on a number of
examples, including high-dimensional data generated from Houston crime data and global temperature measurements. 

\begin{figure}[t]
\centering
\begin{overpic}[width=3.5in]{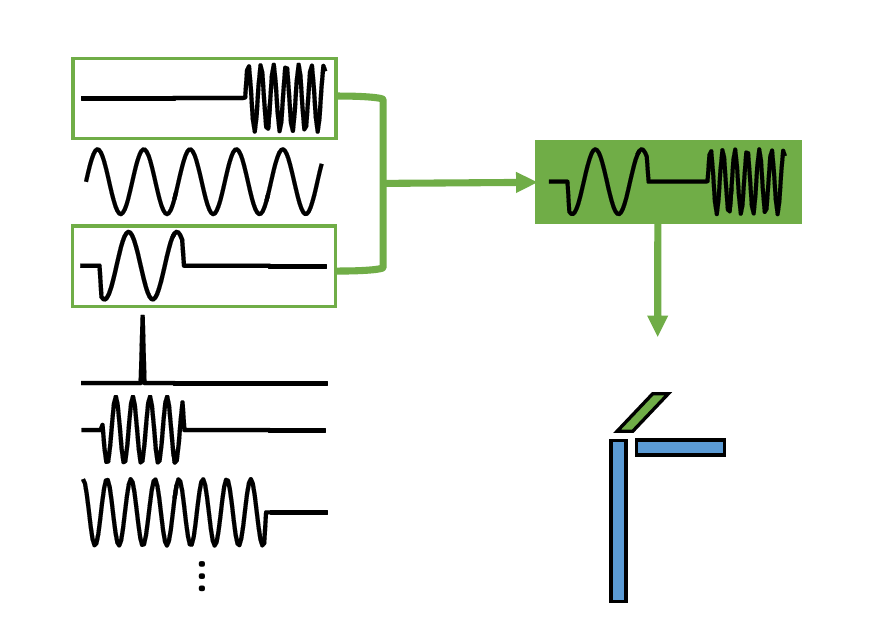}
%\begin{overpic}[width=3.5in,grid,tics=5]{./Figures/Figure4SparseSelection2.pdf}
\put(0,68){{\footnotesize (a) Library of time functions}}
\put(56,58){{\footnotesize (b) Sparse linear combination}}
\put(56,29){{\footnotesize (c) Restricted dimension}}
\put(69,0){{\footnotesize $a_j$}}
\put(80.5,22.6){{\footnotesize $b_j$}}
\put(69,25.5){{\footnotesize $c_j$}}
\put(65,18.5){{\footnotesize $\lambda_j$}}
\end{overpic}
\caption{Sparse selection from an over-complete library. We restrict the third dimension of our CP tensor decomposition (the set of $c_j$ vectors) to be a sparse linear combination of time dynamics functions from a library that we create. (a) We create a library with a variety of functions in time. (b) The algorithm then selects a small (sparse) subset of the library to linearly combine into a $c_j$ vector fitting the time dynamics in the data. (c) This is a restriction on the third dimension of each component.}
\label{fig:SparseSelection}
\end{figure}

\begin{figure}[t]
\centering
\includegraphics[width=3.5in]{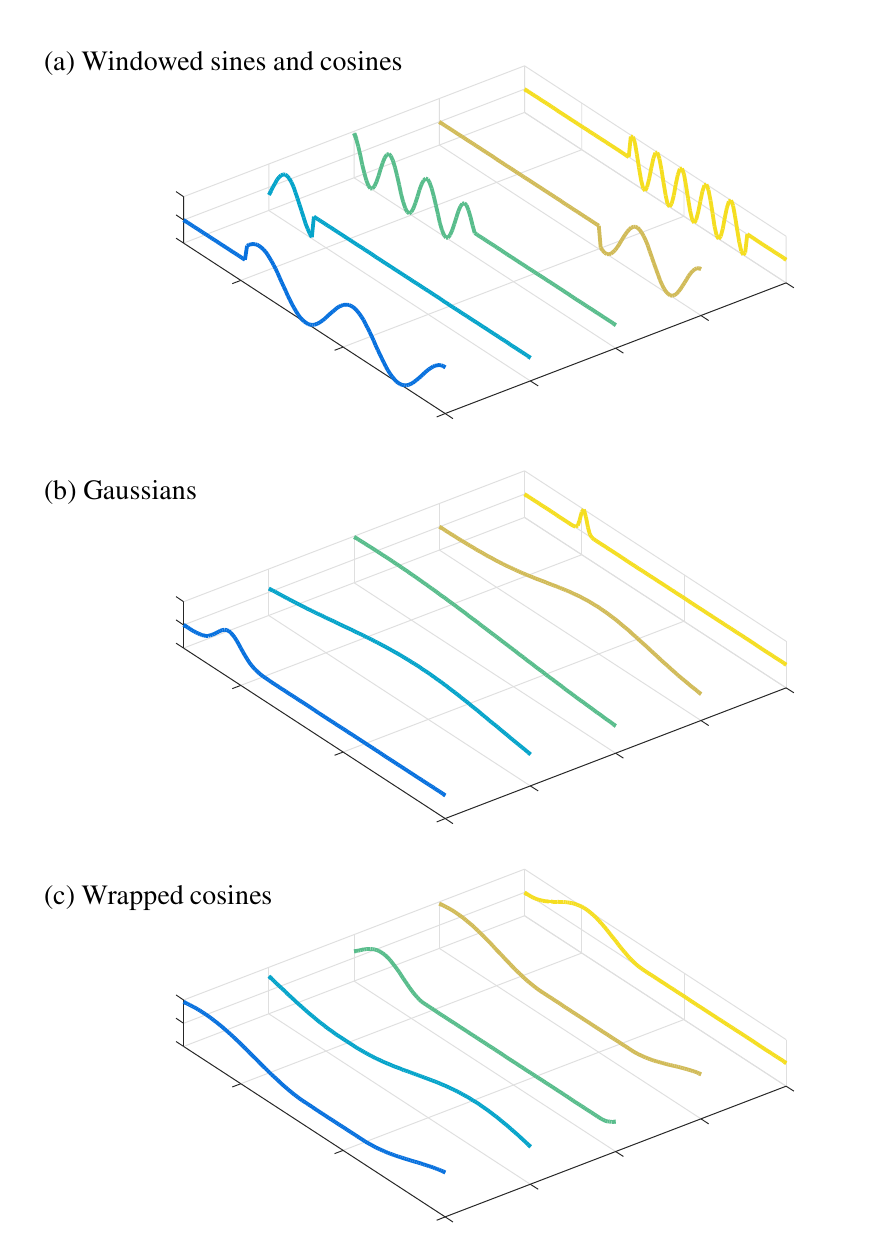}
\caption{Constructing a library. Based on the application, we choose a library of possible time dynamics functions. Options include: (a) Windowed Sines and Cosines. We generate a range of sines and cosines, varying the frequency, width of the window, and center of the window. (b) Gaussians. We fill the library with Gaussian functions, varying the $\mu$ and $\sigma$ parameters. (c) Wrapped Cosines. One way to generate a library that is Gaussian-like but has a period that is the length of the interval is to use one period of a shifted cosine. The frequency and shift can be varied.}
\label{fig:Library}
\end{figure}

The rest of the paper is organized as follows:  Sec.~II develops the basic mathematical 
architecture of the CP tensor decomposition and our refinement, the SCTD algorithm.  This is followed by Sec.~III in which we discuss practical details such as how to select tuning parameters and how to construct an appropriate over-complete library.  %Regularization of the selection algorithm and its relation to the Bayesian information criteria (BIC) is discussed in Sec.~V. 
Sec.~IV tests the algorithm on simulated data, and Sec.~V provides examples demonstrating the
effectiveness of the algorithm on real-world data sets.  Conclusions and an outlook for the SCTD
algorithm are discussed in Sec.~VI. For full details, all MATLAB and R codes used for this paper are available online at 
\url{github.com/BethanyL/SCTD}.

%
%\begin{itemize}
%\item Matrix decompositions are enablers across data science\dots 
%\item Looking for low-rank structure\dots
%\item Frame for spatio-temporal data\dots expect clear signatures\dots
%\item Want interpretability\dots  
%\item We wish to decompose data arrays into components that capture transient dynamics. This is a refinement on foreground / background separation as we are segmenting the foreground into temporally distinct components.
%\item Our contribution is a solution to a Goldilocks problem. On the one hand we have MR DMD (Fourier representation). On the other we have (completely unsupervised) CP tensor factorization.
%\item Illustrative toy example. Show CP Decomposition, MR DMD, and our method (Do we have a name?).
%\item Might briefly discuss DMD in background
%\end{itemize}

\section{Methodology}

In this manuscript, we present a number of modifications to the standard CP tensor decomposition
that are intuitively appealing and improve interpretability of the low-rank modes extracted
from data.  Specifically, we introduce an over-complete library of temporal responses that
constrains the time mode dynamics.  A sparsity-promoting algorithm further selects a small
number of these modes to represent the data.   Thus the procedure can be thought of as 
a sparsity-promoting, constrained optimization problem.

The SCTD method is illustrated at a high level in Figs.~\ref{fig:SparseSelection} and \ref{fig:Library}.  
Fig.~\ref{fig:SparseSelection} shows the selection process whereby a small number
of modes from an over-complete library of temporal functions are selected to best represent the
temporal evolution of the data. We rely on an $\ell_1$ optimization
procedure so as to obtain a sparse representation of the temporal dynamics.  The
algorithm thus restricts the temporal mode in the decomposition.   While the temporal functions populating the library may be arbitrary, the power
of the SCTD relies on the fact that temporal dynamics are typically far from arbitrary.   Fig.~\ref{fig:Library} illustrates some temporal functions
that serve as prototypes that characterize real-world temporal dynamics.  Note that some of these functions are ideally suited for handling
transient dynamics.  Indeed, the success of the method is directly related to 
the temporal library functions included in the regression procedure.

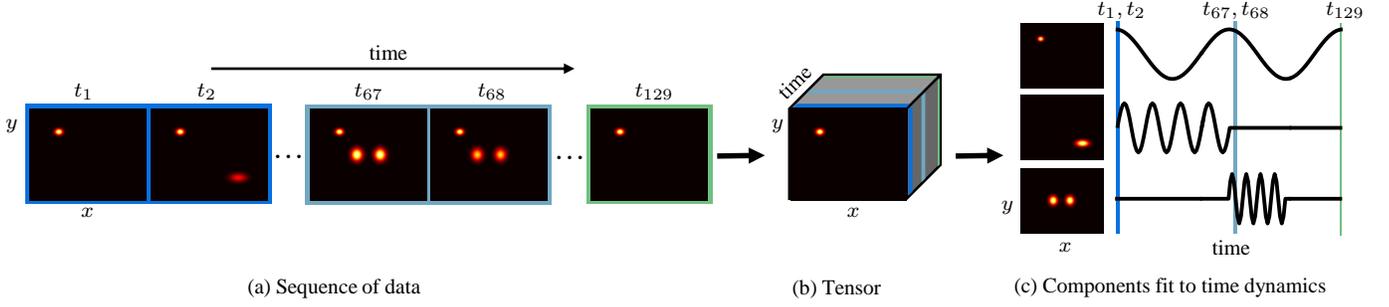
\begin{figure*}[t]
%\centering
\begin{overpic}[width=7.16in]{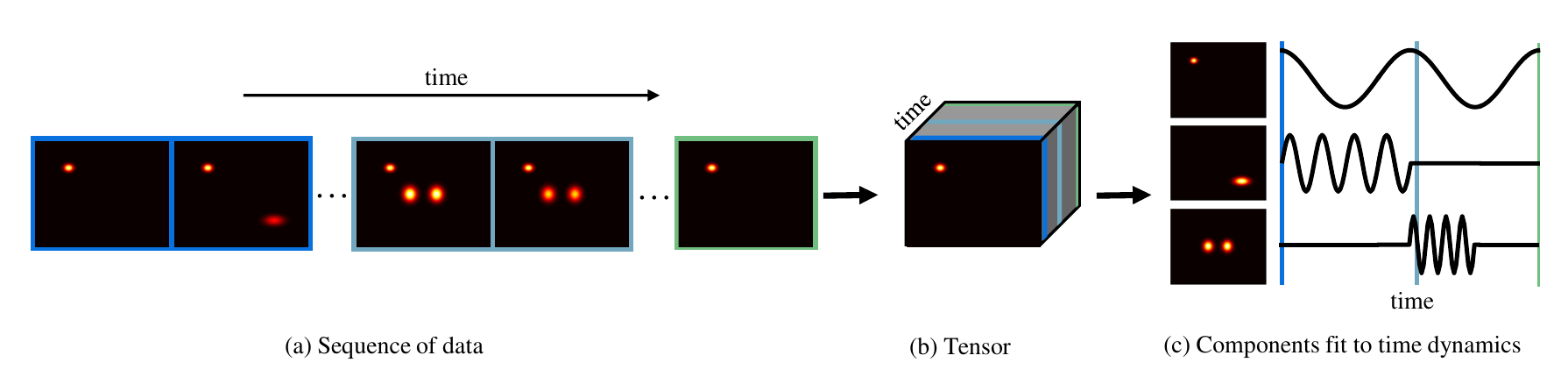}
%\begin{overpic}[width=7.16in,grid,tics=3]{./Figures/Figure1Goal4.pdf}
\put(5.5,16.3){{\footnotesize $t_1$}}
\put(14.5,16.3){{\footnotesize $t_2$}}
\put(26,16.3){{\footnotesize $t_{67}$}}
\put(35,16.3){{\footnotesize $t_{68}$}}
\put(46.5,16.3){{\footnotesize $t_{129}$}}
%\put(24,19){{\footnotesize time}} %(27,19)
\put(6,7.5){{\footnotesize $x$}}
\put(.5,14){{\footnotesize $y$}}
\put(62,7.5){{\footnotesize $x$}}
\put(56.5,14){{\footnotesize $y$}}
\put(77.5,5){{\footnotesize $x$}}
\put(73.3,8){{\footnotesize $y$}}
\put(80.3,22.3){{\footnotesize $t_1, t_2$}}
\put(88,22.3){{\footnotesize $t_{67}, t_{68}$}}
\put(97,22.3){{\footnotesize $t_{129}$}}
\end{overpic}
\caption{Extracting patterns from spatio-temporal data. (a) We begin with a data set where spatial information is collected over time. If we collect two-dimensional data at each time step, we may informally think of the data as a sequence of ``frames.'' (b) The sequence of frames can be saved as a tensor (one data cube) where the third dimension is time. (c) Our goal is to decompose that tensor into a sum of important frame components where each frame component has its own time dynamics. In this example, we see the three components coming in and out of the frames as time passes. The color coding demonstrates how the sample frames in part (a) are combinations of the components shown in part (c).}
\label{fig:Overview}
\end{figure*}

A specific demonstration of the SCTD is shown in Fig.~\ref{fig:Overview}.  In this example,
three different spatial mode structures are combined with three specific time dynamics.
The imposed time dynamics are representative of simple functional forms that are often
difficult for the standard CP or DMD methods to model or resolve, i.e. temporal responses that
have finite time windows of activity.  In this example, the sequence of data snapshots are
gathered into a 3-way data tensor $\T{M}$.  Different snapshots of the dynamics depict
the spatial structure arising from the combination of the different modal structures.  The
objective of the SCTD is to solve the inverse problem:  Given the data tensor $\T{M}$, 
find the low-rank decomposition that correctly reconstructs the spatial modes and their
time dynamics.   The algorithm proposed here, which is based on the CP decomposition,
can indeed recover the three modes and their time dynamics as shown in Fig.~\ref{fig:Overview}.

In the subsections that follow, the technical details of the CP tensor decomposition algorithm
are considered along with strategies for building an over-complete library and enforcing a
parsimonious combination of temporal dynamics prototypes.

%In this note, we propose a CANDECOMP/PARAFAC (CP) \cite{KolBad2009} decomposition in which a mode is sparsely coded. The alternative is motivated by difficulties that dynamic mode decomposition \cite{TuRowLuc2014} has with recovering transient temporal modes.

\subsection{CP Tensor Decompositions}

We begin by first reviewing some useful notation. We denote the $r$th column of a matrix $\M{A}$ by $\V{a}_r$. 
Given matrices $\M{A} \in \Real^{I \times K}$ and
$\M{B} \in \Real^{J \times K}$, their Khatri-Rao product is denoted by $\M{A} \Khat \M{B}$ and is defined to be the $IJ \times K$ matrix of column-wise Kronecker products, namely
\begin{eqnarray*}
\M{A} \Khat \M{B} & = & \begin{pmatrix}
\V{a}_1 \Kron \V{b}_1 & \cdots & \V{a}_K \Kron \V{b}_K
\end{pmatrix}.
\end{eqnarray*}
For an $N$-way tensor $\T{A}$ of size $I_1 \times I_2 \times \cdots \times I_N$, we denote its $\V{i} = (i_1, i_2, \ldots, i_N)$ entry by $\TE{a}{i}$.
The inner product between two $N$-way tensors $\T{A}$ and $\T{B}$ of compatible dimensions is given by
\begin{eqnarray*}
\left \langle \T{A}, \T{B} \right \rangle & = & \sum_{\V{i}} \TE{A}{i}\TE{B}{i}.
\end{eqnarray*}
The Frobenius norm of a tensor $\T{A}$,
denoted by $\lVert \T{A} \rVert_{\text{F}}$, is the square root of the inner product of $\T{A}$ with itself, namely
$\lVert \T{A} \rVert_{\text{F}} = \sqrt{\left \langle \T{A}, \T{A} \right \rangle}$. Finally, the mode-$n$ matricization or unfolding of a tensor $\T{A}$ is denoted by $\Mz{A}{n}$.

%The Frobenius norm of a tensor $\T{X}$, denoted $\lVert \T{X} \rVert_{\text{F}}$ is the natural extension of the Frobenius norm of a matrix, namely
%\begin{eqnarray*}
%\lVert \T{X} \rVert_{\text{F}} & \equiv & \sqrt{\sum_{\bf i} \TE{X}{i}^2},
%\end{eqnarray*}

Let $\T{M}$ represent an $N$-way data tensor of size $I_1 \times
I_2 \times \cdots \times I_N$. We are interested in an $R$-component
 CANDECOMP/PARAFAC (CP) \cite{CaCh70,Ha70} factor model 
\begin{eqnarray}
  \label{eq:cp}
  \T{M} & = & \sum_{r=1}^R \lambda_r \;
  \MnC{A}{1}{r} \circ \cdots \circ \MnC{A}{N}{r},
\end{eqnarray}
where $\circ$ represents outer product and $\MnC{A}{n}{r}$ represents the $r$th column of the 
\emph{factor matrix} $\Mn{A}{n}$ of size $I_n \times R$.  We refer to each
summand as a \emph{component}. Assuming each factor matrix has been
column-normalized to have unit Euclidean length, we refer to the
$\lambda_r$'s as \emph{weights}. We will use the shorthand notation $\T{M} =\llbracket{\V{\lambda}; \Mn{A}{1},\dots,\Mn{A}{N}}\rrbracket$, where $\V{\lambda} = (\lambda_1, \ldots, \lambda_R)\Tra$ \cite{BaKo07}. A tensor that has a CP decomposition is sometimes referred to as a Kruskal tensor.

For the rest of this article, we consider a $3$-way tensor where two modes index state variation and the third mode indexes time variation. 
\begin{eqnarray*}
  \T{M} & = & \sum_{r=1}^R \lambda_r \; \V{A}_{r} \circ \V{B}_{r} \circ \V{C}_{r}.
\end{eqnarray*}
Let $\M{A} \in \Real^{I_1 \times R}$ and $\M{B} \in \Real^{I_2 \times R}$ denote the factor matrices corresponding to the two state modes and $\M{C} \in \Real^{I_3 \times R}$ denote the factor matrix corresponding to the time mode.  This $3$-way decomposition is illustrated in Fig.~\ref{fig:TensorDecomp}.

%We seek the CP decomposition $\T{M}$ that is closest to the data tensor $\T{X}$, namely
%\begin{eqnarray*}
%\min \mathcal{L}(\T{M},\T{X}) \quad \text{such that} \quad \T{M} = \KT{\V{\lambda}; \M{A},\M{B},\M{C}},
%\end{eqnarray*}
%where $\mathcal{L}$ quantifies the discrepancy between $\T{M}$ and $\T{X}$. The canonical discrepancy is the least squares criterion,
%\begin{eqnarray*}
%\mathcal{L}(\T{M}, \T{X}) & = & \lVert \T{M} - \T{X} \rVert_\text{F}^2.
%\end{eqnarray*}
%Note that minimizing the least squares criterion is equivalent to the following maximization problem if $\lVert \T{M} \rVert_{\text{F}} \leq 1$.
%\begin{eqnarray*}
%\max \left \langle \T{M}, \T{X} \right \rangle \quad \text{such that} \quad \text{$\T{M} = \KT{\V{\lambda}; \M{A},\M{B},\M{C}}$ and $\lVert \T{M} \rVert_{\text{F}} \leq 1$}.
%\end{eqnarray*}

\subsection{Sparse Representations in Over-Complete Libraries}

We can further impose structure on the factors of the low-rank decomposition. For example, we could impose sparsity and smoothness on factors \cite{Witten2009, Allen2012}. Here we assume that the time mode can be coded as a sparse linear combination from a known over-complete library $\M{D} \in \Real^{I_3 \times P}$, namely
\begin{eqnarray*}
\M{C} & = & \M{D}\M{Z},
\end{eqnarray*}
where the elements of $\M{Z} \in \Real^{P \times R}$ are predominantly zero, i.e.\@ $\mathop{\rm nnz}(\M{Z}) \ll PR$, and $I_3 \ll P$. This set up can be thought of as a sparse version of CANDELINC (canonical decomposition with linear constraints) \cite{DouglasCarroll1980}.   Figs.~\ref{fig:SparseSelection} and \ref{fig:Library} show both
the constrained decomposition and some example library functions used.

We seek the CP model that maximizes a penalized correlation with the data tensor $\T{X}$ 
 \begin{eqnarray}
\label{eq:log-likelihood}
\That{M} = \underset{\T{M}}{\arg\max}\; f(\T{M}) \equiv \left \langle \T{X} , \T{M} \right \rangle - \tau \lVert \M{Z} \rVert_1 
 \end{eqnarray}
 such that
 \begin{eqnarray*}
  \T{M} & = & \KT{\V{\lambda}; \M{A},\M{B},\M{C}}, \\
   \M{C} & = & \M{D}\M{Z}, \\
 \lVert \V{\lambda} \rVert_2 & \leq & 1, \\
 \lVert \V{a}_r \rVert_2, \lVert \V{b}_r \rVert_2, \lVert \V{z}_r \rVert_2 & \leq & 1 \qtext{for $r = 1, \ldots, R$}.
 \end{eqnarray*}
The matrix norm $\lVert \M{Z} \rVert_1$ is the sum of all the absolute values of $\M{Z}$ and not the induced matrix 1-norm. Thus, the non-negative parameter $\tau$ trades off the degree of correlation of the model to the data and the sparsity level in the loadings $\M{Z}$. The inequality constraints are added to ensure that the feasible set of the optimization problem is compact. The Bolzano-Weierstrass theorem ensures that a solution to the problem exists. Without these constraints, the optimization problem is not well posed as there is no global maximum. %For example, we can construct a sequence $\T{M}^{(n)} = 

We pause to clarify the relationship between the above problem and the sparse coding or dictionary learning problem \cite{Olshausen1997}. Note that we can rewrite the optimization problem in (\ref{eq:log-likelihood}) as
 \begin{eqnarray*}
\text{maximize}\; \left \langle \Mz{X}{3} , (\M{B}\Khat\M{A})\M{\Lambda}\M{D}\Tra\M{Z}\Tra \right \rangle - \tau \lVert \M{Z} \rVert_1 
 \end{eqnarray*}
 such that
 \begin{eqnarray*}
 \lVert \V{\lambda} \rVert_2 & \leq & 1, \\
 \lVert \V{a}_r \rVert_2, \lVert \V{b}_r \rVert_2, \lVert \V{z}_r \rVert_2 & \leq & 1 \qtext{for $r = 1, \ldots, R$.}
 \end{eqnarray*}
If we add the additional constraint $\lVert (\M{B}\Khat\M{A})\M{\Lambda}\M{D}\Tra\M{Z}\Tra \rVert_{\text{F}} = c$ for some constant $c$, then maximizing the penalized correlation is equivalent to minimizing the penalized squared error
\begin{eqnarray*}
\frac{1}{2}\lVert \Mz{X}{3} - \M{W}\M{Z}\Tra \rVert^2_{\text{F}} + \tau \lVert \M{Z} \rVert_1,
\end{eqnarray*}
where $\M{W} = (\M{B}\Khat\M{A})\M{\Lambda}\M{D}\Tra$. Thus, we see that the optimization problem given in (\ref{eq:log-likelihood}) is closely related to a special case of the sparse coding problem where we seek to learn sparse coefficients $\M{Z}$ as well as a dictionary matrix $\M{W}$ which must obey rather strong structural constraints.

\subsection{Algorithm}

We now describe an algorithm for computing our structured low-rank approximation $\T{M}$. Note that the CP constraint $\T{M} = \KT{\V{\lambda}; \M{A},\M{B},\M{C}}$ renders the optimization problem in (\ref{eq:log-likelihood}) non-convex. Note, however, that if we fix all but one of the block variables $\M{A}, \M{B}, \M{Z},$ or $\V{\lambda}$, the optimization problem involves a straightforward concave optimization whose solutions can be written in closed form. Thus, we propose a block coordinate ascent (BCA) algorithm.

One complication of adopting a BCA algorithm is that the updates for $\M{A}, \M{B},$ and $\M{Z}$ each separate into $R$ identical optimization problems. 
To be explicit, consider the problem of updating $\M{Z}$ when $\V{\lambda}, \M{A},$ and $\M{B}$ are fixed.

 \begin{eqnarray}
\max \sum_{r=1}^R \left [\lambda_r  \left \langle \T{X} , \V{a}_r \circ \V{b}_r \circ \M{D}\V{z}_r \right \rangle - \tau \lVert \V{Z}_r \rVert_1  \right ]
 \end{eqnarray}
 such that
 \begin{eqnarray*}
 \lVert \V{z}_r \rVert_2 & \leq & 1 \qtext{for $r = 1, \ldots, R$.}
 \end{eqnarray*}
Consequently, solving for the entire factor matrix $\M{Z}$ at once will yield identical columns, namely $\V{z}_1 = \V{z}_2 = \ldots = \V{z}_R$.
To deal with this degeneracy, we construct $\T{M}$ via deflation. The idea is to find the most correlated rank-1 tensor and then subtract it from the data tensor. We then repeat the procedure on the residual tensor.

We are now ready to summarize at a high level the BCA algorithm. Suppose we have completed $r-1$ rounds so far and let $\T{Y}_r$ denote the residual, namely $\T{X} - \sum_{r'=1}^{r-1} \lambda_r\V{a}_{r'} \circ \V{b}_{r'} \circ \M{D}\V{z}_{r'}$. At the $r$th round, we solve the following optimization problem.

 \begin{eqnarray}
\label{eq:deflate}
\That{M} = \underset{\T{M}}{\arg\max}\; f(\T{M}) \equiv \left \langle \T{Y}_r , \T{M} \right \rangle - \tau \lVert \V{Z}_r \rVert_1 
 \end{eqnarray}
 such that
 \begin{eqnarray*}
  \T{M} & = & \V{A}_r \circ \V{B}_r \circ \V{C}_r, \\
   \V{C}_r & = & \M{D}\V{Z}_r, \\
 \lVert \V{a}_r \rVert_2, \lVert \V{b}_r \rVert_2, \lVert \V{z}_r \rVert_2 & \leq & 1. \\
 \end{eqnarray*}
 
 We again solve the above maximization problem with block coordinate ascent. Once BCA has converged, we determine $\lambda_r$ by solving the problem:
 
 \begin{eqnarray*}
\lambda_r & = & \underset{\lambda}{\arg\min}\; \lVert \T{Y}_r - \lambda \T{M} \rVert_2.
 \end{eqnarray*}
 
 The solution to the above scalar optimization problem is given by
 
 \begin{eqnarray*}
 \lambda_r & = & \frac{\left \langle \T{Y}_r, \T{M} \right \rangle}{\lVert \T{M} \rVert_{\text{F}}^2}.
 \end{eqnarray*}
 
 We now detail the updates for the factors $\V{a}_r, \V{b}_r,$ and $\V{z}_r$.
 
{\bf Updating $\V{a}_r$: }

\begin{eqnarray}
\Vn{A}{n+1}_r %& = & \underset{\lVert \V{A} \rVert_2 \leq 1}{\arg\max}\; \left \langle \Mz{Y}{1}, 
%\V{A}(\Vn{C}{n} \Kron \Vn{B}{n})\Tra \right \rangle \\
& = & \underset{\lVert \V{A} \rVert_2 \leq 1}{\arg\max}\; \left \langle \V{u}_r,\V{A} \right \rangle,
\end{eqnarray}
where $\V{U}_r = \Mz{Y}{1}(\Vn{C}{n}_r \Kron \Vn{B}{n}_r)$. The update simply requires normalizing $\V{u}_r$.
\begin{eqnarray*}
\Vn{A}{n+1}_r & = & \frac{\V{U}_r}{\lVert \V{U}_r \rVert_2}.
\end{eqnarray*}

{\bf Updating $\V{b}_r$: }

\begin{eqnarray}
\Vn{B}{n+1}_r% & = & \underset{\lVert \V{B}_r \rVert_2 \leq 1}{\arg\max}\; \left \langle \Mz{Y}{2}, 
%\V{B}(\Vn{C}{n} \Kron \Vn{A}{n+1})\Tra \right \rangle \\
& = & \underset{\lVert \V{B} \rVert_2 \leq 1}{\arg\max}\; \left \langle \V{V}_r,\V{B} \right \rangle,
\end{eqnarray}
where $\V{V}_r = \Mz{Y}{2}(\Vn{C}{n}_r \Kron \Vn{A}{n+1}_r)$.  The update simply requires normalizing $\V{v}_r$.
\begin{eqnarray*}
\Vn{B}{n+1}_r & = & \frac{\V{V}_r}{\lVert \V{V}_r \rVert_2}.
\end{eqnarray*}

{\bf Updating $\V{z}_r$: }

\begin{eqnarray}
\label{eq:update_z}
\Vn{Z}{n+1}_r %& = & \underset{\lVert \V{Z}_r \rVert_2 \leq 1}{\arg\max}\; \left \langle \Mz{Y}{3}, 
%\M{D}\V{Z}(\Vn{B}{n+1} \Kron \Vn{A}{n+1})\Tra \right \rangle - \tau \lVert \V{z} \rVert_1 \\
& = & \underset{\lVert \V{Z} \rVert_2 \leq 1}{\arg\max}\; \left \langle \V{F}_r ,  
\V{Z} \right \rangle - \tau \lVert \V{Z} \rVert_1,
\end{eqnarray}
where $\V{F}_r = \M{D}\Tra\Mz{Y}{3}(\Vn{B}{n+1}_r \Kron \Vn{A}{n+1}_r)$.

The optimization problem posed in (\ref{eq:update_z}) is a modified lasso problem and has appeared in similar settings \cite{Allen2012,Witten2009,Chi2013}. The update is given by
\begin{eqnarray*}
\Vtilde{z}_r & = & \underset{\V{z}}{\arg\min}\; \frac{1}{2} \lVert \V{f}_r - \V{z} \rVert_2^2 + \tau \lVert \V{z} \rVert_1 \\
\left [\Vtilde{z}_r \right ]_i & = & \sign{\left [\V{f}_r \right ]_i}\max\{\lvert \left [\V{f}_r \right ]_i \rvert - \tau, 0\} \\
\Vn{z}{n+1}_r & = & \begin{cases}
\frac{\Vtilde{z}_r }{\lVert \Vtilde{z}_r \rVert_2} & \text{if $\lVert \Vtilde{z}_r \rVert_2 \not = 0$} \\
\V{0} & \text{otherwise}.
\end{cases}
\end{eqnarray*}

The derivation of this update rule is given in the Appendix. We then update $\lambda_r$, followed by calculating the next residual $\T{Y}_{r+1} = \T{Y}_r - \lambda_r \V{a}_r \circ \V{b}_r \circ \M{D}\V{z}_r$.
%We have the following convergence guarantees for the BCA algorithm.
%
%%\begin{proposition}
%{\bf Proposition}: 
%The limit points of the sequence of BCA iterates are stationary points to the optimization problem.
%%\end{proposition}
%
%A proof is given in the Appendix.

%Note that the $\lambda_r$ was not included in the updates for $\V{a}_r$ and $\V{b}_r$ but is included for the update for $\V{z}_r$. In the two former cases, $\lambda_r$ falls out of during the normalization step. In the latter case, it needs to be included prior to the soft-thresholding.

\section{Details of SCTD Algorithm}

Now that we are familiar with the basic procedure, we next discuss important details in the SCTD. We first describe how the sparsity-inducing parameter $\tau$ is chosen. We then describe how the over-complete library is constructed.

\subsection{Picking The Regularization Parameter}
% Genevera / BIC
At each iteration, we use the Bayesian Information Criterion (BIC) \cite{Schwarz1978} to pick regularization parameter $\tau_r$. The BIC is a quantitative score that balances how well the model fits the data against how complicated the model is. In the context of the SCTD, a constrained rank-1 Kruskal tensor with low BIC corresponds to a rank-1 Kruskal tensor which fits the data well in light of how many free parameters were used in fitting it. As defined in \cite{Allen2012}, for this problem, the BIC criterion is
\[
\mbox{BIC}(\tau_r) \!=\! \log \! \left[ \frac{\| \T{Y}_r \!\!-\! \lambda_r\V{a}_{r} \!\circ\! \V{b}_{r} \!\circ\! \M{D}\V{z}_{r} \|_F^2}{I_1 I_2 I_3} \right] + \frac{\log(I_1 I_2 I_3)}{I_1 I_2 I_3} |\{\V{z}_{r}\}|,
\]
where $|\{\V{z}_r\}|$ is the number of non-zero elements of $\V{z}_r$. This can be derived from each update being an $\ell_1$-norm penalized regularization problem. 

%$\T{Y}_r$ denote the residual, namely $\T{X} - \sum_{r'=1}^{r-1} \lambda_r\V{a}_{r'} \circ \V{b}_{r'} \circ \M{D}\V{z}_{r'}$.

We use the BIC criterion to pick the best $\tau_r$ from a range of options. We further refine the value of $\tau_r$ by checking the neighborhood of the current best option until the neighborhood is sufficiently small or the BIC curve is sufficiently constant on that neighborhood. An upper limit of $\tau_r$ is the point at which all entries of $\V{z_r}$ are zero. Since we accumulate small amounts of error on each iteration, we wish to encourage increasing levels of sparsity as $r$ increases. We thus use $\tau_{r-1}$ as a lower bound for $\tau_r$, unless $\tau_{r-1}$ is greater than the current upper bound.

\subsection{Constructing the Over-Complete Library}
We can choose an over-complete library based on knowledge of the application area or data set.  Some natural candidates are displayed in Fig.~\ref{fig:Library}. If we expect periodic but transient dynamics, we may choose to populate the library with windowed sines and cosines, varying the frequencies of the sines and cosines and the widths and shifts of the windows. If we anticipate transient phenomena that are non-periodic, it may be appropriate to include Gaussians with a range of means and variances. If the time domain itself is periodic, such as hour of the day, then we might improve the results by including dynamics that have this period. For example, to allow a Gaussian-like mode to vary smoothly through the night, we could generate cosines with varying frequencies and shifts, but only include one period of the cosine (see ``wrapped cosines'' in Fig.~\ref{fig:Library}).

\begin{figure*}[!t]
\centering
\begin{overpic}[width=7.16in]{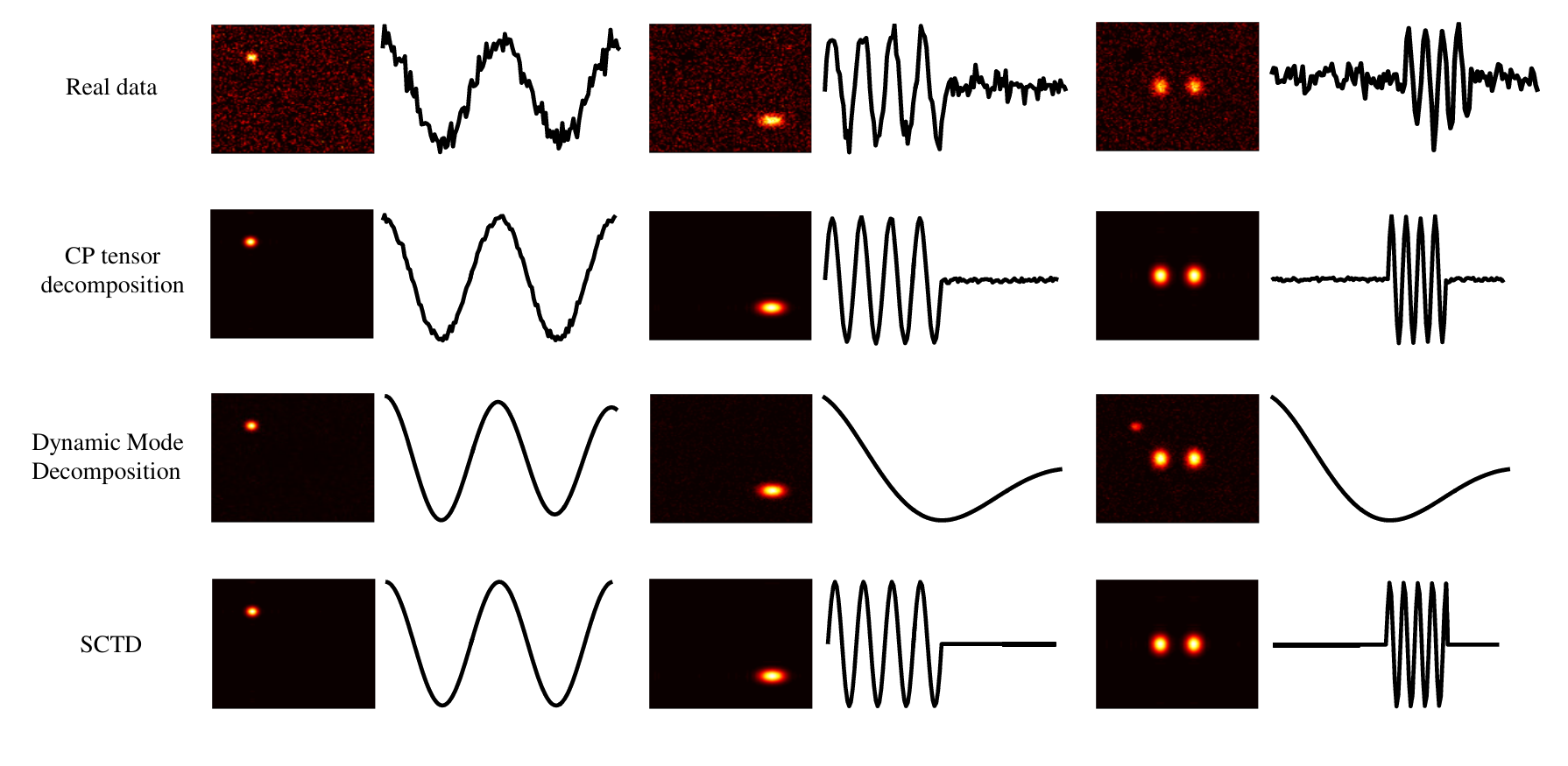}
%\begin{overpic}[width=7.16in,grid,tics=2]{./Figures/Figure2ComparisonMethods2.pdf}
\put(26,13){{\footnotesize $e^{(-.002+.101i)t}$}}
\put(55,13){{\footnotesize $e^{(-.013+.045i)t}$}}
\put(84,13){{\footnotesize $e^{(-.013-.045i)t}$}}
\put(24.5,1.5){{\footnotesize $\cos .098t, t \in [0,128]$}}
\put(52.5,1.5){{\footnotesize $\sin .392t, t \in [0,63.5]$}}
\put(81,1.5){{\footnotesize $\sin .783t, t \in [64.8,98.1]$}}
\end{overpic}
\caption{Comparing methods on a simulated data set. The data set, a 3-way tensor, is generated as described in Fig.~\ref{fig:Overview}, except that noise is added. We hope that a method can decompose the tensor into its three noiseless components. A traditional CP tensor decomposition sometimes falls into a good local minimum and decomposes the data correctly. Clean spatial modes are found, but some noise in the time dynamics is maintained. The time dynamics are not fit to analytic expressions. The Dynamic Mode Decomposition tries to fit clean time dynamics functions to the spatial modes. However, it is restricted to Fourier modes and cannot handle the windowed behavior in this data set. It also does not correctly separate the third spatial mode. The SCTD finds clean spatial modes and fits smooth time dynamics to each component. The output includes the exact functions that were fit to the time dynamics.}
\label{fig:InitialComparison}
\end{figure*}

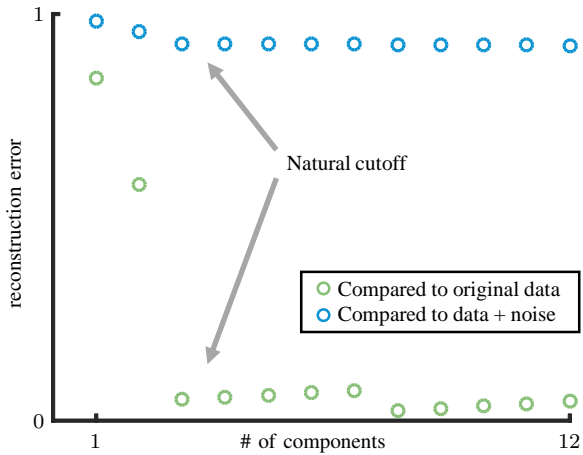
\begin{figure}[!t]
\centering
\begin{overpic}[width=3.5in]{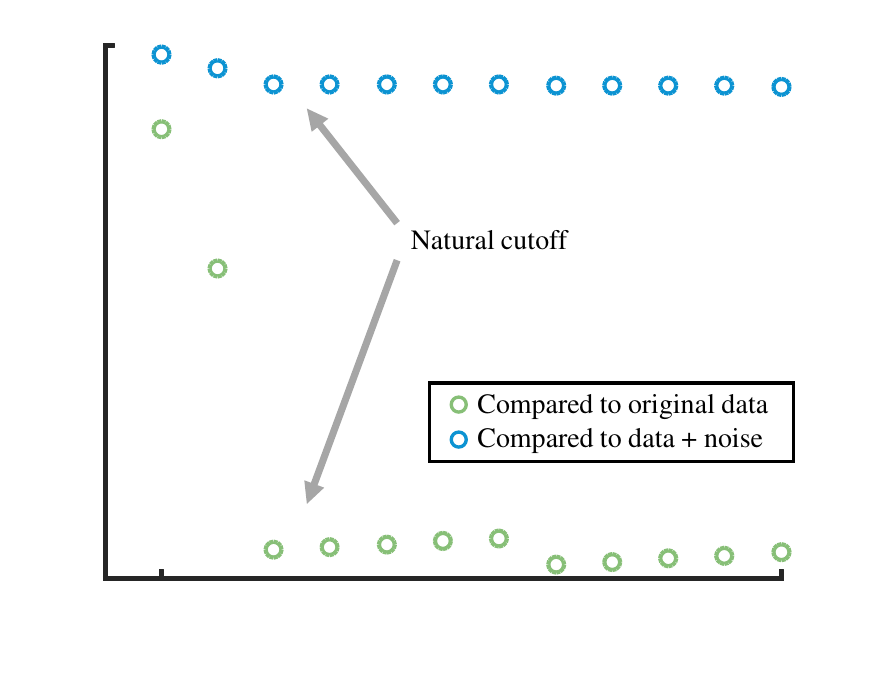}
%\begin{overpic}[width=3.5in,grid,tics=5]{./Figures/Figure6ReconError.pdf}
\put(17.5,7){{\footnotesize $1$}}
\put(87.5,7){{\footnotesize $12$}}
\put(40,7){{\footnotesize \# of components}}
\put(5,28){{\footnotesize \rotatebox{90}{reconstruction error}}}
\put(9,10){{\footnotesize $0$}}
\put(9,71){{\footnotesize $1$}}
%\put(43.5,44.5){{\footnotesize natural cutoff}}
\end{overpic}
\caption{Reconstruction error curve. We can choose the number of components to keep in the SCTD by considering the trade-off between error and complexity. Here we see diminishing returns in reconstruction error after the inclusion of the first three components, suggesting that a rank-3 approximation sufficiently captures the majority of systematic variation in the data. We calculate the error in two ways---by comparing the reconstruction to the original clean data and to the noisy data.}
\label{fig:ReconErrorCurve}
\end{figure}

\section{Simulation Experiments}

We begin by testing the SCTD on a simulated data set similar to Fig.~\ref{fig:Overview}. Recall that this data set is composed of three spatio-temporal modes (specifically, it is a Kruskal tensor with rank three). We can think of this data set as a video or sequence of frames. Our goal is to decompose it into three modes (a rank-three Kruskal tensor) with an analytical description for the temporal dimension.  Although the SCTD is exceptional for the data in Fig.~\ref{fig:Overview},  the example
is limited since no noise was included in the data.  

We next consider a more realistic example shown 
in Fig.~\ref{fig:InitialComparison}.  In this experiment, we added white Gaussian noise with standard deviation $\sigma$ in the frequency domain to the data $\left( u_n(t) = \mathcal{F}^{-1}[\hat{u}(\omega) + \sigma \mathcal{N}(0,1)] \right)$.  In this case, we used $\sigma = 3$, resulting in a signal-to-noise ratio of 0.1374, where signal-to-noise ratio is defined as the ratio of the summed squared magnitude of the signal to the summed squared magnitude of the noise. The algorithm outlined in Sec.~II can now be applied
to the data and a direct comparison can be made to a CP decomposition and a DMD reduction. In particular, for the CP decomposition, we use the CP\_ALS function in the Matlab Tensor Toolbox \cite{TensorToolbox}, \cite{TTB_Dense}, which uses an alternating least squares algorithm.
Fig.~\ref{fig:InitialComparison} shows that despite the inclusion of noise, the modes and temporal
dynamics can be cleanly extracted using the SCTD. Indeed, analytic forms for the time dynamics
can be discovered.  In comparison, the CP algorithm gives a decomposition with noisy time modes which lack analytic description.  The DMD algorithm (using data flattening) can give analytic expressions for
the time dynamics, but the temporal expressions are significantly flawed due to the fact that DMD
cannot handle such transient and/or intermittent time dynamics, i.e. only time dynamics of the form $\exp (\omega t)$ 
are allowed. 

The SCTD also provides a diagnostic for performing an $r$-rank truncation.
For an SVD decomposition, the singular values provide the requisite metric for truncation.
%Similarly, the $\lambda_j$ values provide the necessary information in the SCTD for truncation.   
Similarly, Fig.~\ref{fig:ReconErrorCurve} shows the decay of reconstruction error as
a function of the number of tensor modes. We can choose the rank, or number of components to keep in the SCTD, by considering the trade-off between error and complexity. Here we see diminishing returns in reconstruction error after the inclusion of the first three components, suggesting that a rank-3 approximation sufficiently captures the majority of the systematic variation in the data.
% not using the lambda_j values here for truncation - only looking at reconstruction error 

\begin{figure*}[!t]
\centering
\begin{overpic}[width=7.16in]{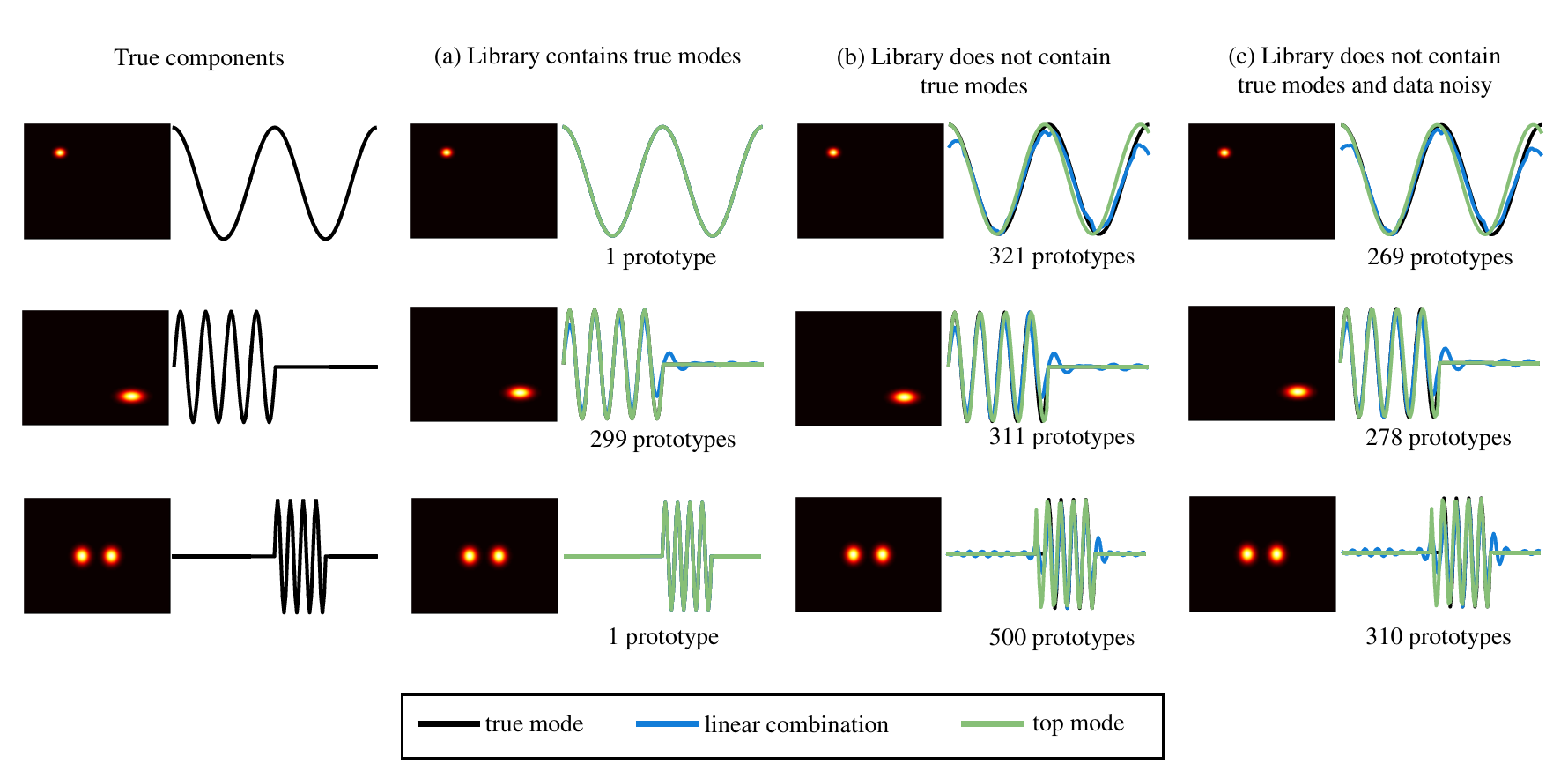}
\end{overpic}
\caption{Results on simulated data set. We repeat for reference the three true components that compose the data set. (a) When the library contains the correct time dynamics functions, the SCTD does a good job of recovering them. (b) When the library does not contain the exact right modes, the SCTD uses more prototypes to fit the data, but still chooses a sparse number. (c) When we additionally make the data noisy, the SCTD is robust. It chooses more prototypes, but if an especially simple output is desired, using just the prototype with the highest coefficient is accurate. See more detail in Tab.~\ref{tab:TableResults}.}
\label{fig:SimResults}
\end{figure*}

To further explore the example shown in Fig.~\ref{fig:InitialComparison}, we consider
a number of different cases which highlight the use of the algorithm and the choice
of library prototypes.  Thus we consider the following:
 
\begin{itemize}
\item \emph{Case (a): Library contains true modes.} We start with an easy case. We construct a library with 3000 prototypes, including the true temporal modes, and we do not add noise to the data. We see in Fig.~\ref{fig:SimResults} and Table~\ref{tab:TableResults} that in two iterations, the SCTD picks exactly one prototype, and in one iteration, the SCTD picks 299  from the 3000 and accumulates a small amount of error. 
%apply MR DMD, CP Decomposition, our method?

% check that library doesn't happen to include true modes?
\item \emph{Case (b): Library does not contain true modes.} Next, we want to assess how robust the SCTD is to ``model misspecification'': We construct another library of 3000 prototypes but do not ``cheat'' by including the true time dynamics. As we can see in Fig.~\ref{fig:SimResults} and Tab.~\ref{tab:TableResults}, in this experiment, the method uses extra prototypes (about $10\%$ of the library) and accumulates more error. However, the factor accuracy only reduces from 0.989 to 0.948.

%apply MR DMD, CP Decomposition, our method?

\item \emph{Case (c): Library does not contain true modes and the data is noisy.} Finally, we increase the difficulty by  adding white Gaussian noise to the data ($\sigma = 1$). The results are very similar to Case (b) without noise (see Fig.~\ref{fig:SimResults} and Tab.~\ref{tab:TableResults}). Note that although the resulting analytic expression of hundreds of modes is not simple, if you want a simple analytical expression, you can pick the mode with the largest coefficient and still maintain accuracy. The top mode is plotted in green on top of the linear combination (blue) and the true mode (black) in Fig.~\ref{fig:SimResults}. 

%\item %\note[BL]{For all methods, we compare accuracy (Factor match score? like Tamara uses) and timing}
\end{itemize}

\begin{table}[!t]
\renewcommand{\arraystretch}{1.1}
\setlength{\tabcolsep}{5pt}
% if using array.sty, it might be a good idea to tweak the value of
% \extrarowheight as needed to properly center the text within the cells
\caption{Details to accompany Fig.~\ref{fig:SimResults}}
%\caption{Table Accompanying Fig.~\ref{fig:SimResults}. For each experiment, we give the number of elements that the method chose from the library. Then we calculate the error in reconstructing the data set from our tensor decomposition. We display the range of frequencies in the elements chosen from the library. Finally, we give the accuracy of each mode, since we know how the data was generated.}
\label{tab:TableResults}
\centering

\begin{tabular}{|c|c|c|c|c|}
\hline
\multirow{2}{*}{case} & \# prototypes  & relative & factor & frequency of  \\
& chosen & error & accuracy & top mode \\
\hline
\multirow{3}{*}{(a)} & \multirow{3}{*}{301 (3.6\%)} & \multirow{3}{*}{.1144} & \multirow{3}{*}{.988924} & .0982 \\
& & & & .3927 \\
 & & & & .7854 \\
\hline
\multirow{3}{*}{(b)} & \multirow{3}{*}{1132 (13.5\%)} & \multirow{3}{*}{.2329} & \multirow{3}{*}{.948130} & .1026 \\
& & & & .3846 \\
 & & & & .7692 \\
\hline
\multirow{3}{*}{(c)} & \multirow{3}{*}{857 (10.2\%)} & \multirow{3}{*}{.6947} & \multirow{3}{*}{.937874} & .1026 \\
& & & & .3846 \\
& & & & .7692 \\
\hline
\end{tabular}
\end{table}

%\note[BL]{add frequency of top mode to table?}

\begin{figure}
\centering
\includegraphics[width=3.5in]{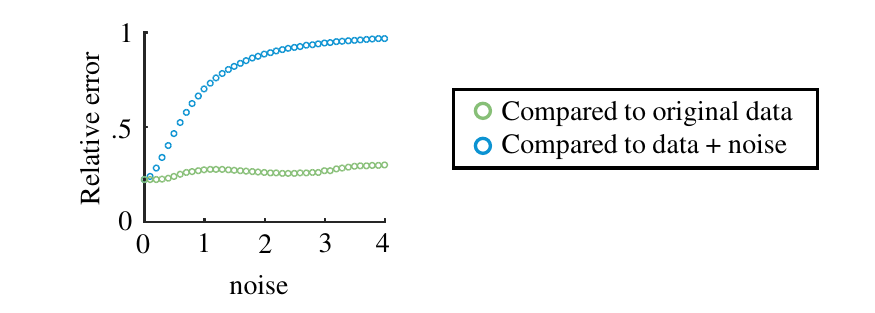}
\caption{Varying the noise. In Figs.~\ref{fig:InitialComparison} and \ref{fig:SimResults} and Tab.~\ref{tab:TableResults}, we displayed results on noisy data. Here we vary the amount of noise to display the robustness of the SCTD. The value of $\sigma$ ranges $0.1\text{--}4$ while the SNR ranges $123.8\text{--}0.101$. As the noise increases, the error in the reconstruction of the original data increases. Note that the cases of $\sigma = 3$ and $\sigma = 1$ are displayed in Figs.~\ref{fig:InitialComparison} and \ref{fig:SimResults}, respectively. The increase in error is slow when the error is in terms of the noiseless data.}
\label{fig:Noise}
\end{figure}

\begin{figure}
\centering
\includegraphics[width=3.5in]{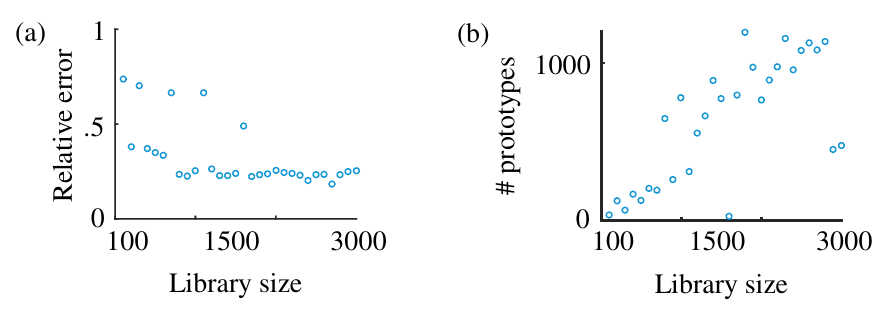}
\caption{Varying the library size. In Fig.~\ref{fig:SimResults} and Tab.~\ref{tab:TableResults}, we displayed results on a library with 3,000 prototypes. Here we vary the size of the library to consider the tradeoffs. Once we have a reasonably large library, the relative error is consistent. However, the number of selected prototypes roughly grows with the library size. Thus to limit complexity, we may wish to pick a library size that is sufficient for low error reconstructions but is not larger than necessary.}
\label{fig:DictSize}
\end{figure}

Next, we consider Case (c) in more detail. So far, we have seen two examples of this case. In Fig.~\ref{fig:InitialComparison}, the library contains  40,000 prototypes and the white noise has standard deviation $\sigma = 3$. In Fig.~\ref{fig:SimResults} (and Tab.~\ref{tab:TableResults}), the library contains 3,000 prototypes and the white noise has standard deviation $\sigma = 1$. We now consider the effect of $\sigma$ (Fig.~\ref{fig:Noise}) and the effect of the library size (Fig.~\ref{fig:DictSize}). 

In Fig.~\ref{fig:Noise}, we fix the library size to 50,000 and vary $\sigma$. As the magnitude of the noise increases, so does the error. However, this growth in error is slow when the error is measured against the original (noiseless) data.  For context, see Fig.~\ref{fig:InitialComparison} for a visualization of data with $\sigma = 3$. 

%\note[BL]{switch to signal-to-noise ratio.}

%\note[BL]{make experiments more consistent so comparable here?}

%\note[BL]{Add inset plots visualizing noise to Figure 8?}

In Fig.~\ref{fig:DictSize}, we consider data that has no noise and vary the library size. Once the library is sufficiently large, the relative error does not improve. However, the number of prototypes chosen grows. %This is an opportunity to make a bias-variance trade off.

%\note[BL]{address another form of accuracy, like factor score.}

\section{Real Data Examples}
We now apply the SCTD to two real-world data sets exhibiting complex spatio-temporal dynamics with intermittency.  These
examples illustrate the power of the SCTD to produce interpretable 
results, especially in the constrained time dynamics. Figures were rendered with the ggmap and ggplot2 R packages \cite{ggmap, Wickham2009}.

\subsection{Houston Crime}
Data mining is beginning to be applied to a myriad of law enforcement problems \cite{McCue2014}. These techniques can be used to help agencies deploy their employees more efficiently, predict the outcomes of new initiatives, and identify trends in crime in order to take preventative measures. 

We apply the SCTD to a data set, collected by the Houston Police Department, with 85,622 crimes occurring in Houston from January to August 2010. We use a preprocessed version of the data included in the ggmap R package \cite{ggmap}. We create a 3-way tensor of counts of these crimes. The dimensions are type of crime (aggravated assault, auto theft, burglary, robbery, or theft), crime beat (118 options), and hour of day (0--23). We then apply the SCTD to this data set using a mix of the three types of functions displayed in Fig.~\ref{fig:Library}---windowed sines and cosines, Gaussians, and wrapped cosines. 

\begin{figure*}[t]
\centering
\includegraphics[width=7.16in]{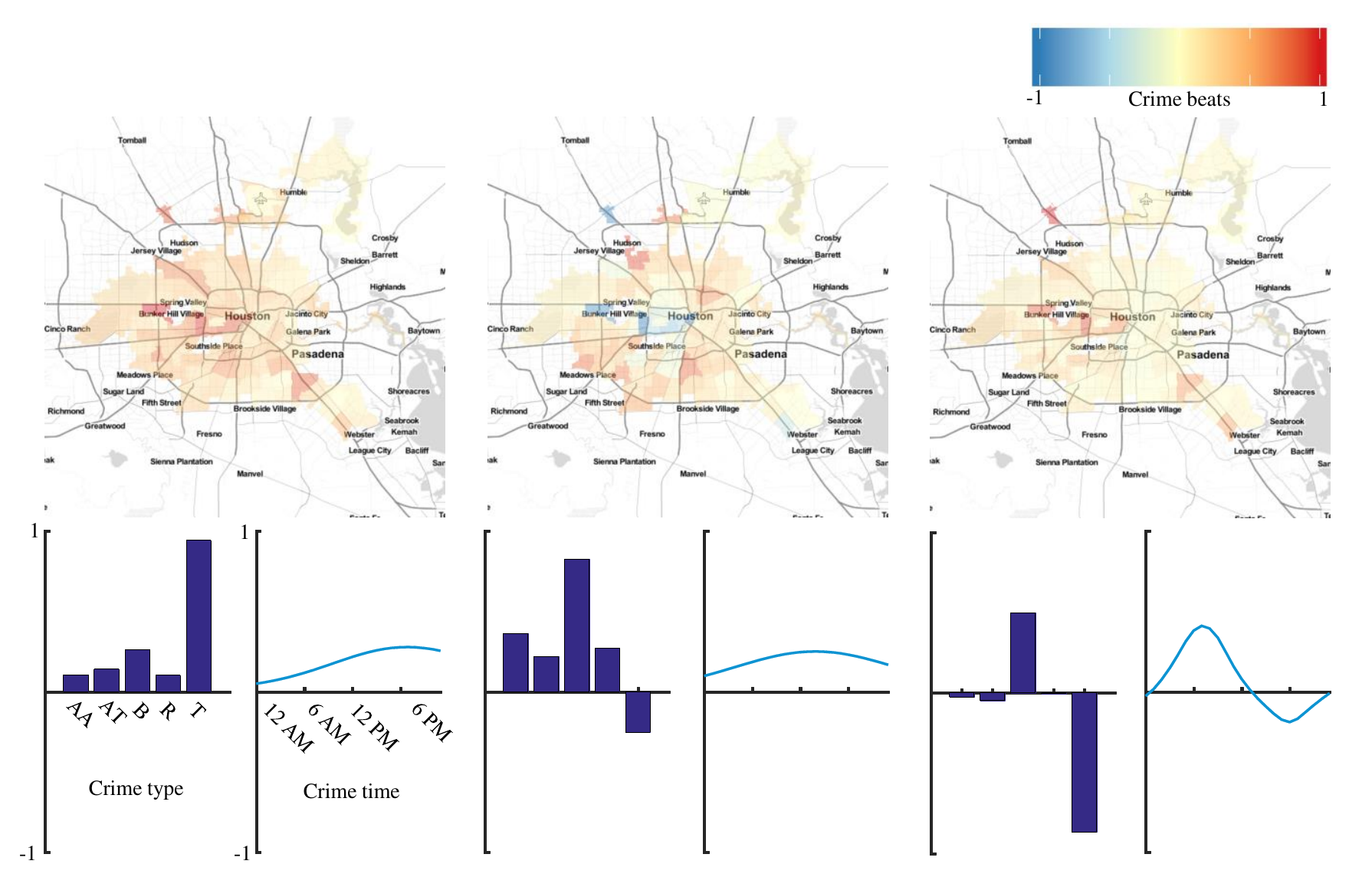}
\caption{Results on Houston crime data set using SCTD. We start with a data set of Houston crime where the first dimension is type of crime, the second is crime beat, and the third is hour of the day (0--23). The five crimes considered are aggravated assault (AA), auto theft (AT), burglary (B), robbery (R), and theft (T). We decompose the data set with the SCTD and display the first three modes here. Our method finds sets of beats behaving similarly and assigns smooth, interpretable time dynamics.}%\add[BL]{Comment on some patterns}}
\label{fig:HoustonOurs}
\end{figure*}

The first three components of the SCTD are displayed in Fig.~\ref{fig:HoustonOurs}.  In the first component, most beats are at least lightly included, although some are more intense. Theft in the evening is especially emphasized. The second component adds non-theft crimes to a different set of hot spots and subtracts non-theft crime from some of the beats that were important in the first component. The third component re-emphasizes some of the same beats, this time adding burglary and subtracting theft in the morning and conversely adding theft and subtracting burglary in the evening. 

\begin{figure*}
\centering
\includegraphics[width=7.16in]{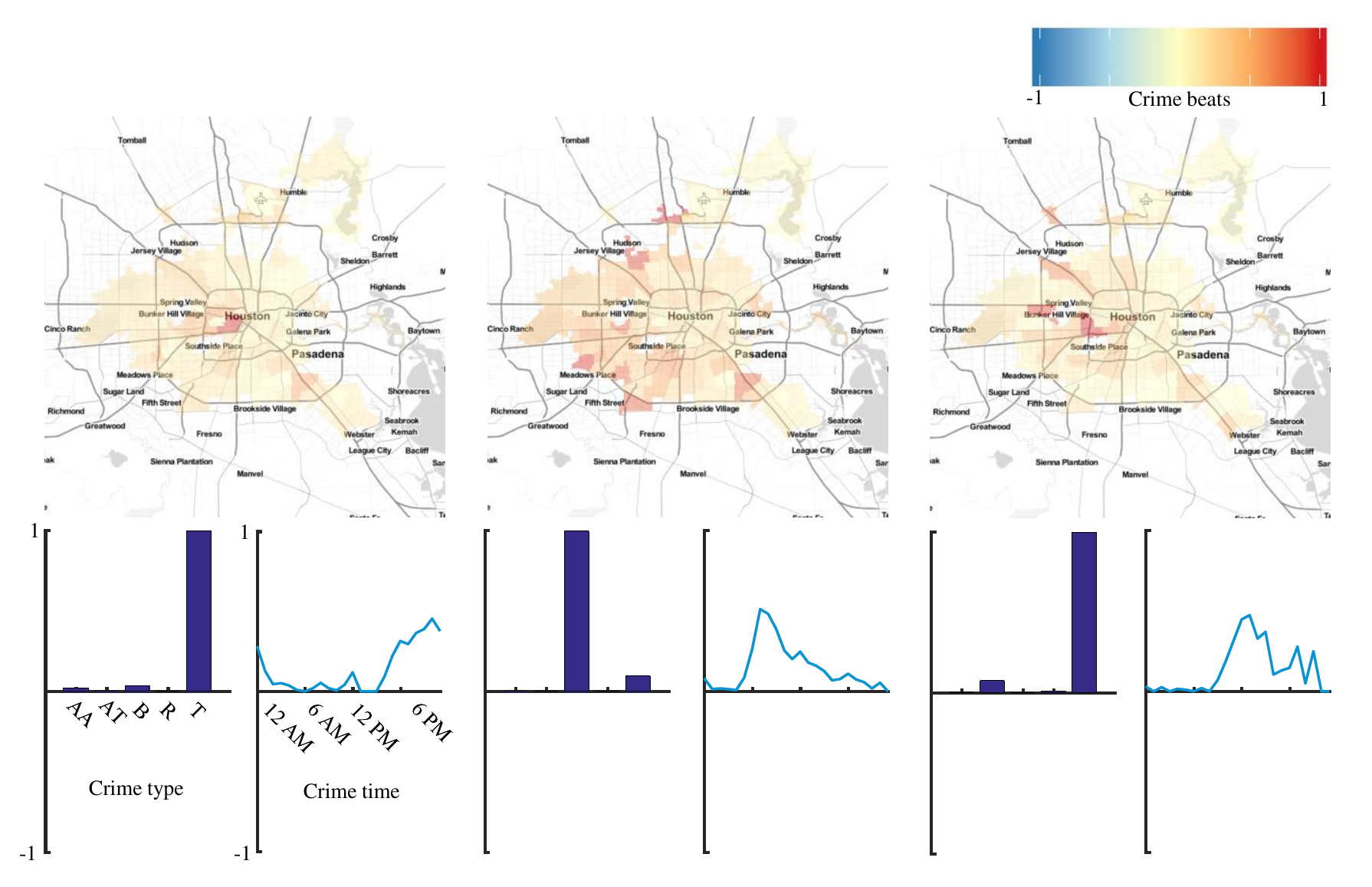}
\caption{Results on Houston crime data set using CP\_APR. We decompose the Houston crime data set again, but this time with the with the CP\_APR tensor decomposition for comparison. We display the first three modes here. Note that the time dynamics are noisy. }%\add[BL]{Comment on some patterns}}
\label{fig:HoustonCP_APR}
\end{figure*}

In Fig.~\ref{fig:HoustonCP_APR}, we compare our results to the first three components from CP\_APR (the non-negative CP tensor decomposition with alternating Poisson regression \cite{CPAPR} as implemented in the Matlab Tensor Toolbox \cite{TensorToolbox}). The results using the SCTD are much smoother and more interpretable in the time dimension, but many of the same beats are considered important.  Again, the ability to produce analytically tractable time functions helps our ability to both interpret and predict the
dynamics associated with this complex, socially-inspired network.
%\note[BL]{HoustonCrime.mat vs. HoustonCrime2.mat}\\

%\note[BL]{what to call CP\_APR method?}

\subsection{El Ni\~{n}o}
Sensor and imaging technologies (oceanic, terrestial and satelite) have led to a significant increases in 
climate data and a limited but growing understanding of how to extract meaningful information from it. Interest in this interdisciplinary field has spawned, for example, the annual International Workshop on Climate Informatics \cite{ClimateInformatics}.

\begin{figure}
\centering
\includegraphics[width=3.5in]{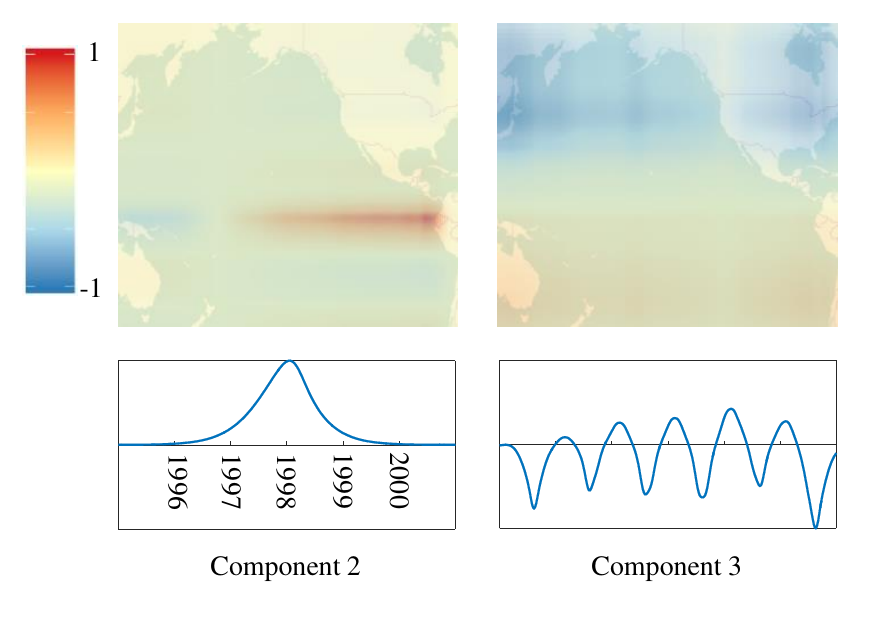}
\caption{Results on ocean surface temperature data set. We start with a data set of ocean surface temperature over time. The dimensions are longitude, latitude, and time. Here we display a sample of the components found by the SCTD. The second component finds the El Ni\~{n}o event of 1997--1998, a warm band in the central and east-central equatorial Pacific. The third contain annual variation, split over the equator. }
\label{fig:ElNino}
\end{figure}

We demonstrate the SCTD on a data set of sea surface temperatures. The data are freely available from the NOAA/OAR/ESRL PSD, Boulder, Colorado, USA. We used the weekly sea surface temperature from the NOAA\_OI\_SST\_V2 data set, which can be downloaded from \url{http://www.esrl.noaa.gov/psd/}. In particular, we consider the Pacific Ocean from 1995 through the end of 2000. We subtracted the background from the data using DMD \cite{DMDbackground}, and then we created a library with  a combination of Gaussians and windowed sines and cosines. Two of the first twelve modes that the SCTD extracted are shown in Fig.~\ref{fig:ElNino}. The second component finds one-time phenomena related to the El Ni\~{n}o event of 1997--1998. In particular, we see unusually warm temperatures in the eastern Pacific ocean, especially near Peru, but almost stretching to New Guinea. By mid-to-late 1997, unusually cool waters occurred near the coast of Australia.  The third component finds annual variation in temperature, split over the equator. 

%\note[BL]{official citation and more careful El Nino description}\\
%\note[BL]{more clear labels of years (beginning vs. end)}
% add citations for more R packages? and maps 

\begin{figure}
\centering
\includegraphics[width=3.5in]{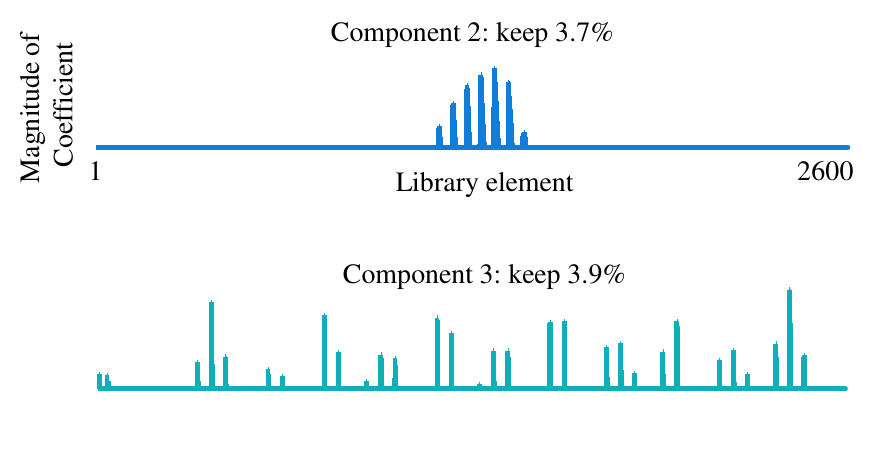}
\caption{Results on ocean surface temperature data set, continued. This figure gives further information about the results in Fig.~\ref{fig:ElNino}. We demonstrate the sparsity of the time dynamics by plotting the magnitudes of the coefficients in each $\V{z}_r$. }
\label{fig:ElNinoBars}
\end{figure}

These results could not be obtained with standard DMD because the El Ni\~{n}o event is not a Fourier mode.  However, recent 
innovations around multi-resolution analysis and DMD (the multi-resolution DMD algorithm~\cite{mrDMD}) does allow for a significantly
improved description.  Likewise, a traditional CP tensor decomposition might extract similar patterns, they would not be accompanied with a sparse analytic description. We show that the SCTD choses a sparse linear combination of our over-complete library in Fig.~\ref{fig:ElNinoBars}. 

% add more details about exact parameters, etc. 

%\subsection{Bucket}
%\begin{figure}
%\centering
%\includegraphics[width=3.5in]{./Figures/Figure12BucketResults.jpg}
%\caption{Results on Oscillating Paint Can Video. We start with a video of a paint can oscillating. A generic CP tensor decomposition finds the paint can in its first term but the time dynamics are noisy and do not capture the correct frequency. DMD also finds the paint can, but not until the ninth term. The earlier terms are decomposing noise in the data. It finds the correct time dynamics for the paint can but because the spatial modes are unrestricted, the spatial mode is noisy. Finally, our method finds the paint can on the first term. The spatial mode is simple and the time dynamics extract the correct frequency.}
%\label{Bucket}
%\end{figure}

%\note[BL]{ possibly add Bucket back in}

%\IEEEpubidadjcol

\section{Conclusion}

Data-driven discovery has become ubiquitous across the sciences, leading to
the rise of the fourth paradigm of scientific discovery~\cite{fourth}.  Critical in
meeting the challenges of this emerging paradigm is the development of
algorithms that are capable of extracting meaningful and interpretable low-dimensional features
from data that is high-dimensional and includes many distinct dimensions.
The success of machine learning is largely due to its ability to represent 
data in low-dimensional feature spaces where data can be more effectively analyzed, classified 
and clustered.   Matrix decomposition techniques, which project to low-rank subspaces
via some underlying optimization algorithm, are the workhorses of the data science
industry.  For instance, Principal Component Analysis is now standard across almost
every field of the engineering, social, biological and physical sciences.  This SVD-based method provides a least-square fitting algorithm for data, thus
providing low-rank subspaces that best represent the features of the data.

The success of the SVD is difficult to overestimate.  It is simply the most
dominant and successful matrix decomposition method being used today.  The
SVD requires, however, that multi-dimensional data first be {\em flattened} before being
processed through the decomposition.   This can lead to less parsimonious 
fitting than if the data was preserved in its original $N$-way data tensor.
Tensor decompositions, on the other hand, allow the data to be preserved
in its original multi-dimensional context, which is especially advantageous for
categorical data.  Although tensors have been the subject
of active research for the past four decades, it has been difficult for tensor
decompositions to displace standard SVD with flattening decompositions.
This is in part due to the multitude of potential tensor decompositions available
to the practitioner, i.e. it is not unique.  Moreover, the SVD has numerous
enhancements for handling high-dimensional data, such as the randomized 
SVD~\cite{martinsson,ben_erichson}, which enables efficient computation of the matrix decomposition even
with extraordinarily large data.

In this manuscript, we have developed what we think is a highly useful innovation
to the standard CP tensor decomposition.  By constraining the time dimension of
the tensor decomposition, a more intuitively appealing and interpretable decomposition
can be achieved.  Indeed, analytic solution forms for the time dependency of the
data decomposition can be extracted.  This is done by using an over-complete library
of potential temporal functions in order to select the best candidate functions via
sparse regression.  This work merges three distinct mathematical methods:  tensor
decompositions, sparse regression, and over-complete libraries.  The success of the
SCTD method is demonstrated on a number of simulated problems and two real-world applications
where preserving the tensor nature of the data is highly desirable and advantageous.
The SCTD method provides a viable data-discovery algorithm that can be used
in a host of settings where low-rank features of an $N$-way data tensor need
to be analyzed.  It should also be noted that one can easily envision also constraining other
dimensions of the data, not just the time dimension.

Ultimately, the most useful data analysis techniques developed allow for interpretable diagnostics 
which are also predictive in nature.  The SCTD advances a theoretical framework for tensor decompositions 
that provides an intuitively appealing framework for understanding the rich time dynamics of low-rank decompositions without
requiring data-flattening.  With the emergence of many categorical data structures, this can be especially
appealing.  Thus, we render a tensor decomposition package that is user-friendly and aids in identifying important
dynamics structures in data, including intermittent phenomena, which are very difficult for standard
tensor, DMD and PCA-like methods to deduce.

\section*{Appendix}

\subsection*{Notation Details}
\label{sec:notation-details}

%{\bf Khatri-Rao product:} Give two matrices $\M{A}$ and $\M{B}$ of sizes $I_1 \times R$ and $I_2
%\times R$, then $\M{C} = \M{A} \Khat \M{B}$ is a matrix of size $I_1
%I_2 \times R$ such that
%\begin{displaymath}
%  \M{C} =
%  \begin{bmatrix}
%    \MC{A}{1} \Kron \MC{B}{1}
%    & \MC{A}{2} \Kron \MC{B}{2}
%    & \cdots
%    & \MC{A}{R} \Kron \MC{B}{R}
%  \end{bmatrix},
%\end{displaymath}
%where the Kronecker product of two vectors of size $I_1$ and $I_2$ is
%a vector of length $I_1I_2$ given by
%\begin{displaymath}
%  \V{a} \Kron \V{b} =
%  \begin{bmatrix}
%    a_1 \V{b} \\ a_2 \V{b} \\ \vdots \\ a_{I_1} \V{b}
%  \end{bmatrix}.
%\end{displaymath}

{\bf Matricization of a tensor:} The mode-$n$ matricization or unfolding of a tensor
$\T{A}$ is denoted by $\Mz{A}{n}$ and is of size $I_n \times J_n$
where $J_n \equiv \prod_{m \neq n} I_m$. In this case, the tensor element with index 
$\MI{i}$ maps to matrix element $(i,j)$ where
\begin{displaymath}
  i = i_n \qtext{ and }
  j = 1 + \displaystyle\sum_{{k=1}\atop{k \neq n}}^N (i_k - 1) \left( \prod_{{m=1}\atop{m \neq
        n}}^{k-1} I_m \right).
\end{displaymath}

\subsection*{Derivation of $\V{z}$ update:}

We prove that the update for $\V{z}$ is the solution to the optimization problem given in (\ref{eq:update_z}).

\begin{proof}
The negative of the Lagrangian for (\ref{eq:update_z}) (to express the optimization as a minimization) is given by
\begin{eqnarray*}
\mathcal{L}(\V{z},\gamma) & = & -\langle \V{f}, \V{z} \rangle + \tau \lVert \V{Z} \rVert_1 + \gamma(\lVert \V{z} \rVert_2^2 - 1).
\end{eqnarray*}
The KKT conditions are given by
\begin{eqnarray*}
\V{f} - 2\gamma\V{z} & \in & \tau \partial \lVert \V{z} \rVert_1 \\
\lVert \V{z} \rVert_2^2 & \leq & 1 \\
\gamma & \geq & 0 \\
\gamma (\lVert \V{z} \rVert_2^2 - 1) & = & 0.
\end{eqnarray*}
There are two cases to consider. If $\Vtilde{z} = \V{0}$, then the pair $(\V{z},\gamma) = (\V{0},0)$ satisfies the KKT conditions. The stationarity condition is satisfied since $\V{f} \in \tau \partial \lVert \V{0} \rVert_1$ since $\Vtilde{z} = \V{0}$ solves the lasso problem. The other conditions are easily verified.

If $\Vtilde{z} \not = \V{0}$, then the pair $(\V{z},\gamma) = (\Vtilde{z}/\lVert \Vtilde{z} \rVert_2,0)$ satisfies the KKT conditions. The only tricky condition to verify is the stationarity condition. The other conditions are easy to verify. Since $\Vtilde{z}$ is the solution to the lasso problem we have that
\begin{eqnarray*}
\V{f} - \Vtilde{z} & \in & \tau \partial \lVert \Vtilde{z} \rVert_1 \\
\V{f} - 2 \left (\frac{\lVert \Vtilde{z} \rVert_2}{2} \right )\frac{\Vtilde{z}}{\lVert \Vtilde{z} \rVert_2} & \in &
\tau \partial \left \lVert \lVert \Vtilde{z} \rVert \frac{\Vtilde{z}}{\lVert \Vtilde{z} \rVert_2} \right \rVert_1.
\end{eqnarray*}
Note that we have used the fact that $\partial \lVert \V{z} \rVert_1 = \partial \lVert c \V{z} \rVert_1$ for all $c > 0$. Therefore, the pair $(\V{z},\gamma) = (\Vtilde{z}/\lVert \Vtilde{z} \rVert_2, \lVert \Vtilde{z} \rVert_2/2)$ satisfies the KKT conditions.
\end{proof}

\section*{Acknowledgment}

J. N. Kutz would like to acknowledge support from the Air Force Office of Scientific Research (FA9550-15-1-0385).

\ifCLASSOPTIONcaptionsoff
  \newpage
\fi

\AtEndEnvironment{thebibliography}{

}

\bibliographystyle{unsrt}
\bibliography{TensorPaper}

\begin{thebibliography}{10}

\bibitem{Pearson:1901}
Karl Pearson.
\newblock On lines and planes of closest fit to systems of points in space.
\newblock {\em Philosophical Magazine Series 6}, 2(11):559--572, 1901.

\bibitem{hotellingJEdPsy33_1}
Harold Hotelling.
\newblock Analysis of a complex of statistical variables with principal
  components.
\newblock {\em Journal of Educational Psychology}, 24:417--441, September 1933.

\bibitem{hotellingJEdPsy33_2}
Harold Hotelling.
\newblock Analysis of a complex of statistical variables with principal
  components.
\newblock {\em Journal of Educational Psychology}, 24:498--520, October 1933.

\bibitem{lorenzMITTR56}
Edward~N. Lorenz.
\newblock Empirical orthogonal functions and statistical weather prediction.
\newblock Technical report, Massachusetts Institute of Technology, December
  1956.

\bibitem{Lumley:1970}
John~L. Lumley.
\newblock {\em Stochastic tools in turbulence}.
\newblock Applied mathematics and mechanics. Academic press, New York, London,
  1970.

\bibitem{HLBR_turb}
Philip Holmes, John~L. Lumley, and Gal Berkooz.
\newblock {\em Turbulence, coherent structures, dynamical systems, and
  symmetry}.
\newblock Cambridge monographs on mechanics. Cambridge University Press,
  Cambridge, England, 2nd edition, 2012.

\bibitem{KolBad2009}
Tamara~G. Kolda and Brett~W. Bader.
\newblock Tensor decompositions and applications.
\newblock {\em SIAM Review}, 51(3):455--500, September 2009.

\bibitem{Bishop}
Christopher~M. Bishop.
\newblock {\em Pattern Recognition and Machine Learning (Information Science
  and Statistics)}.
\newblock Springer-Verlag New York, Inc., Secaucus, NJ, USA, 2006.

\bibitem{Murphy}
Kevin~P. Murphy.
\newblock {\em Machine Learning: A Probabilistic Perspective}.
\newblock The MIT Press, 2012.

\bibitem{CaCh70}
J.~Douglas Carroll and Jih-Jie Chang.
\newblock Analysis of individual differences in multidimensional scaling via an
  {N}-way generalization of ``{Eckart-Young}'' decomposition.
\newblock {\em Psychometrika}, 35:283--319, 1970.

\bibitem{Ha70}
Richard~A. Harshman.
\newblock Foundations of the {PARAFAC} procedure: Models and conditions for an
  ``explanatory" multi-modal factor analysis.
\newblock {\em UCLA working papers in phonetics}, 16:1--84, 1970.
\newblock Available at
  \url{http://www.psychology.uwo.ca/faculty/harshman/wpppfac0.pdf}.

\bibitem{tensorDMD}
Stefan Klus, Patrick Gel\ss, Sebastian Peitz, and Christof Sch\"utte.
\newblock Tensor based dynamic mode decomposition.
\newblock arXiv:1512.06527 [math.NA], 2016.

\bibitem{TuRowLuc2014}
Jonathan~H. Tu, Clarence~W. Rowley, Dirk~M. Luchtenburg, Steven~L. Brunton, and
  J.~Nathan Kutz.
\newblock On dynamic mode decomposition: Theory and applications.
\newblock {\em Journal of Computational Dynamics}, 1(2):391--421, 2014.

\bibitem{mrDMD}
J.~Nathan Kutz, Xing Fu, and Steven~L. Brunton.
\newblock Multiresolution dynamic mode decomposition.
\newblock {\em SIAM Journal on Applied Dynamical Systems}, 15(2):713--735,
  2016.

\bibitem{BaKo07}
Brett~W. Bader and Tamara~G. Kolda.
\newblock Efficient {MATLAB} computations with sparse and factored tensors.
\newblock {\em SIAM Journal on Scientific Computing}, 30(1):205--231, December
  2007.

\bibitem{Witten2009}
Daniela~M. Witten, Robert Tibshirani, and Trevor Hastie.
\newblock A penalized matrix decomposition, with applications to sparse
  principal components and canonical correlation analysis.
\newblock {\em Biostatistics}, 10(3):515--534, 2009.

\bibitem{Allen2012}
Genevera Allen.
\newblock Sparse higher-order principal components analysis.
\newblock In {\em Proceedings of the 15th International Conference on
  Artificial Intelligence and Statistics}, 2012.

\bibitem{DouglasCarroll1980}
J.~Douglas~Carroll, Sandra Pruzansky, and Joseph~B. Kruskal.
\newblock Candelinc: A general approach to multidimensional analysis of
  many-way arrays with linear constraints on parameters.
\newblock {\em Psychometrika}, 45(1):3--24, 1980.

\bibitem{Olshausen1997}
Bruno~A. Olshausen and David~J. Field.
\newblock Sparse coding with an overcomplete basis set: A strategy employed by
  v1?
\newblock {\em Vision Research}, 37(23):3311 -- 3325, 1997.

\bibitem{Chi2013}
Eric~C. Chi, Genevera~I. Allen, Hua Zhou, Omid Kohannim, Kenneth Lange, and
  Paul~M. Thompson.
\newblock Imaging genetics via sparse canonical correlation analysis.
\newblock In {\em Biomedical Imaging (ISBI), 2013 IEEE 10th International
  Symposium on}, pages 740--743, 2013.

\bibitem{Schwarz1978}
Gideon Schwarz.
\newblock Estimating the dimension of a model.
\newblock {\em The Annals of Statistics}, 6(2):461--464, 03 1978.

\bibitem{TensorToolbox}
Brett~W. Bader, Tamara~G. Kolda, et~al.
\newblock Matlab tensor toolbox version 2.6.
\newblock Available online, February 2015.

\bibitem{TTB_Dense}
Brett~W. Bader and Tamara~G. Kolda.
\newblock Algorithm 862: {MATLAB} tensor classes for fast algorithm
  prototyping.
\newblock {\em ACM Transactions on Mathematical Software}, 32(4):635--653,
  December 2006.

\bibitem{ggmap}
David Kahle and Hadley Wickham.
\newblock ggmap: Spatial visualization with ggplot2.
\newblock {\em The R Journal}, 5(1):144--161, 2013.

\bibitem{Wickham2009}
Hadley Wickham.
\newblock {\em ggplot2: Elegant Graphics for Data Analysis}.
\newblock Springer-Verlag New York, 2009.

\bibitem{McCue2014}
Colleen McCue.
\newblock {\em Data mining and predictive analysis: Intelligence gathering and
  crime analysis}.
\newblock Butterworth-Heinemann, 2014.

\bibitem{CPAPR}
Eric~C. Chi and Tamara~G. Kolda.
\newblock On tensors, sparsity, and nonnegative factorizations.
\newblock {\em SIAM Journal on Matrix Analysis and Applications},
  33(4):1272--1299, December 2012.

\bibitem{ClimateInformatics}
Valliappa Lakshmanan, Eric Gilleland, Amy McGovern, and Martin Tingley,
  editors.
\newblock {\em Machine Learning and Data Mining Approaches to Climate Science},
  Proceedings of the 4th International Workshop on Climate Informatics.
  Springer, 2015.

\bibitem{DMDbackground}
Jacob Grosek and J~Nathan Kutz.
\newblock Dynamic mode decomposition for real-time background/foreground
  separation in video.
\newblock {\em arXiv preprint arXiv:1404.7592}, 2014.

\bibitem{fourth}
Tony Hey, Stewart Tansley, and Kristin~M. Tolle, editors.
\newblock {\em The Fourth Paradigm: Data-Intensive Scientific Discovery}.
\newblock Microsoft Research, 2009.

\bibitem{martinsson}
Nathan Halko, Per-Gunnar Martinsson, and Joel~A Tropp.
\newblock Finding structure with randomness: Probabilistic algorithms for
  constructing approximate matrix decompositions.
\newblock {\em SIAM review}, 53(2):217--288, 2011.

\bibitem{ben_erichson}
N.~B. Erichson, J.~N. Kutz, S.~L. Brunton, and S.~Voronin.
\newblock Randomized matrix decompositions using r.
\newblock {\em arXiv preprint arXiv:1608.02148}, 2016.

\end{thebibliography}

\end{document}